\documentclass[fleqn,10pt]{wlscirep}
 \pdfoutput=1
\usepackage[utf8]{inputenc}
\usepackage[T1]{fontenc}
\usepackage{lineno}

\usepackage[linesnumbered,ruled,vlined]{algorithm2e}
\usepackage{subcaption}
\usepackage{amsmath}
\SetKwInput{KwInput}{Input}                
\SetKwInput{KwOutput}{Output}              
\usepackage{multirow}

\usepackage{graphicx}
\usepackage{hyperref}
\usepackage[square,sort,comma,numbers]{natbib}
\usepackage{subfiles}

\newcommand{\ts}{\textsuperscript}

\title{A Neuromorphic Dataset for Object Segmentation in Indoor Cluttered Environment}

\author[1,2]{Xiaoqian Huang}
\author[4]{Kachole Sanket}
\author[1]{Abdulla Ayyad}
\author[4]{Fariborz Baghaei Naeini}
\author[4]{Dimitrios Makris}
\author[1,3*]{Yahya Zweiri}
\affil[1]{Advanced Research and Innovation Center (ARIC), Khalifa University, Abu Dhabi, UAE}
\affil[2]{Khalifa University Center for Autonomous Robotic Systems (KUCARS), Khalifa University, Abu Dhabi, UAE} 
\affil[3]{Department of Aerospace Engineering, Khalifa University, Abu Dhabi, UAE}
\affil[4]{School of Computer Science and Mathematics, Kingston University, London, UK}

\affil[*]{corresponding author(s): Yahya Zweiri (yahya.zweiri@ku.ac.ae)}

\begin{abstract}
Taking advantage of an event-based camera, the issues of motion blur, low dynamic range and low time sampling of standard cameras can all be addressed. 
However, there is a lack of event-based datasets dedicated to the benchmarking of segmentation algorithms, especially those that provide depth information which is critical for segmentation in occluded scenes. 
This paper proposes a new Event-based Segmentation Dataset (ESD), a high-quality 3D spatial and temporal dataset for object segmentation in an indoor cluttered environment. 
Our proposed dataset ESD comprises 145 sequences with 14,166 RGB frames that are manually annotated with instance masks. Overall 21.88 million and 20.80 million events from 
two event-based cameras in a stereo-graphic configuration are collected, respectively. 
To the best of our knowledge, this densely annotated and 3D spatial-temporal event-based segmentation benchmark of tabletop objects is the first of its kind. 
By releasing ESD, we expect to provide the community with a challenging segmentation benchmark with high quality. 

\end{abstract}
\begin{document}

\flushbottom
\maketitle

\thispagestyle{empty}

\noindent Please note: Abbreviations should be introduced at the first mention in the main text – no abbreviations lists or tables should be included. The structure of the main text is provided below.

\section*{Background \& Summary}


In the 4\ts{th} industrial revolution, the demand for robots to perform multiple tasks is highly increased. Robots equipped with grippers are increasingly popular and essential for grasping tasks, because they provide the industry with the benefit of cutting manufacturing time while improving throughput. The bulk of these tasks requires the robots to be proficient in gripping objects of different shapes, weights, and textures. However, the majority of present techniques are used to train robots to perform tasks that are suitable for a structured environment with prior knowledge of the scene and objects. Such tasks are prone to high error and are tremendously difficult to fully automate, especially in unstructured environments\cite{Chitta2012MobileExecution}. Because objects in unstructured environments are out-of-order and are of unknown shape and geometry, robotic systems have to perceive and understand the scene using robotic vision rather than relying on prior knowledge and models of objects. Thus, to tackle this issue, robotic perception is key to make robots localize, segment, and grasp objects in an unstructured environment. 

At present, most vision-based applications and research rely on traditional vision sensors such as RGB and RGBD sensors. 
However, traditional frame-based cameras have distinct shortcomings of high power consumption and large storage requirements due to continuous full-frame sensing and storage. Moreover, properties of low sampling rate and motion blur may affect the perceiving quality for many vision-based applications. For instance, due to the low sampling rate of the conventional RGB camera, the fast-moving speed of conveyor belts in production lines introduces motion blur on pictures taken by a standard camera \cite{zhang2019vision}. Thus, the accuracy and success rate 
of object picking and placing are reduced  at the perceiving stage. 
The neuromorphic vision sensor is inspired by biological systems such as fly eyes, which can sense data in parallel and real-time with a micro second-level sampling rate \cite{indiveri2000neuromorphic, lichtsteiner2008128}. 
Building on these unique properties of event cameras, an increasing amount of research is explored based on the neuromorphic vision to avoid motion blur and improve efficiency, such as object tracking \cite{glover2016event}, depth estimation \cite{rebecq2018emvs}, autonomous driving \cite{chen2020event}, and robotic grasping \cite{naeini2019novel, baghaei2020dynamic, muthusamy2021, huang2022real}. 

As a fundamental pre-processing step for these perception-related tasks, segmentation plays a key role in estimating the properties of each object. Especially in vision-based robotic grasping applications, localization and geometric information of each object are required to devise a specific grasping plan \cite{muthusamy2020neuromorphic}. In other words, the quality of perception and segmentation would directly affect the grasping quality. %
In recent years, learning-based approaches to segmentation and other vision-based tasks triggered a massive surge. 
Datasets are significant for computer vision supervised learning methods\cite{EveringhamTheChallenge}. 
Moreover, datasets allow the comparison among various algorithms to provide benchmarks \cite{Deng2010ImageNet:Database}. 
Several RGB and RGBD-based segmentation datasets were constructed to provide ground truth for the training and evaluation of deep-learning-based segmentation approaches. For instance, EasyLabel \cite{Suchi2019EasyLabel:Datasets} offers instance segmentation RGB-D dataset with point-wise labeled point-clouds information for cluttered objects in an indoor environment, where the depth height and the objects in clutter are varied. Also, synthetic dataset TOD was generated for unknown object segmentation \cite{Xie2020UnseenEnvironments}. Besides, there are many other public conventional datasets, such as MSCOCO \cite{LinMicrosoftContext}, PascalVoc\cite{EveringhamTheChallenge}, and CityScape\cite{Cordts2016TheUnderstanding} for multiple tasks including segmentation, object detection, and classification. 
In addition, amounts of conventional vision-based objects segmentation approaches were developed, such as FCN \cite{long2015fully}, U-NET\cite{ronneberger2015u}, and DeepLab\cite{chen2017deeplab} are commonly utilized as evaluation benchmarks.

However, research on event-based segmentation is still in the primary stage of development. Unlike the booming research on the instance segmentation of conventional frame-based vision, little research has been done for event-based instance segmentation of tabletop objects. 
Current solutions for event-based instance segmentation are commonly based on clustering. For example, event-based mean shift clustering approaches were developed in \cite{barranco2018real, huang2022real} using 2D spatial and temporal information, however, they fail to segment occluded objects. Such a limitation may be addressed by using the depth information of RGBD imagery\cite{xia2015situ}. 
Furthermore, events with depth information can provide the ground truth for deep learning-based depth estimation approaches, such as spiking neural networks-based depth estimation from mono event camera \cite{hidalgo2020learning}. 
But there are no existing deep learning-based approaches for neuromorphic instance segmentation of tabletop objects, due to the lack of labeled data for training and testing. 
Instead of developing instance segmentation approaches, transfer learning of semantic segmentation networks can be a possible and quick way to achieve instance segmentation tasks. There are several approaches targeting event-based semantic segmentation for autonomous driving, such as EV-SegNet (2019) \cite{alonso2019ev}, VID2E (2019)\cite{gehrig2020video}, EVDistill (2021)\cite{wang2021evdistill}, EV transfer (2022)\cite{messikommer2022bridging}, and ESS (2022)\cite{sun2022ess}. However, features provided by pure events are limited compared to RGB frames. The cross-modal networks, such as SA-GATE \cite{alonso2019ev} and CMX \cite{liu2022cmx}, are being investigated nowadays to obtain abundant information from both events stream and complementary RGB frames. 

To address this gap, we constructed an Event-based Segmentation Dataset (ESD) of tabletop objects in cluttered scenes, the first of its kind. Particularly, two subsets ESD-1 and ESD-2 are separated as training and testing data for unseen objects tasks.  Events, vision data, and depth are acquired with two \textit{Davis 346c} event cameras and an \textit{Intel D435} RGBD camera attached at the end effector of \textit{UR10} robot. 
The ESD dataset contains events streams and frames from event cameras, raw RGB frames and depth maps from RGBD cameras, moving speed and position of the end effector of UR10 robot. Data were collected under various conditions including different objects, moving speed and trajectory of cameras, lighting conditions, and distance between cameras and tabletop. Events are labeled with depth information and RGBD frames from the conventional camera are also provided in our dataset. 
Moreover, we rigorously evaluate several widely used segmentation methods on our proposed ESD to demonstrate the challenges. 

\section*{Methods}


\subsection*{Experimental setup}
\begin{figure}[ht]
\centering

\includegraphics[width = 5in]{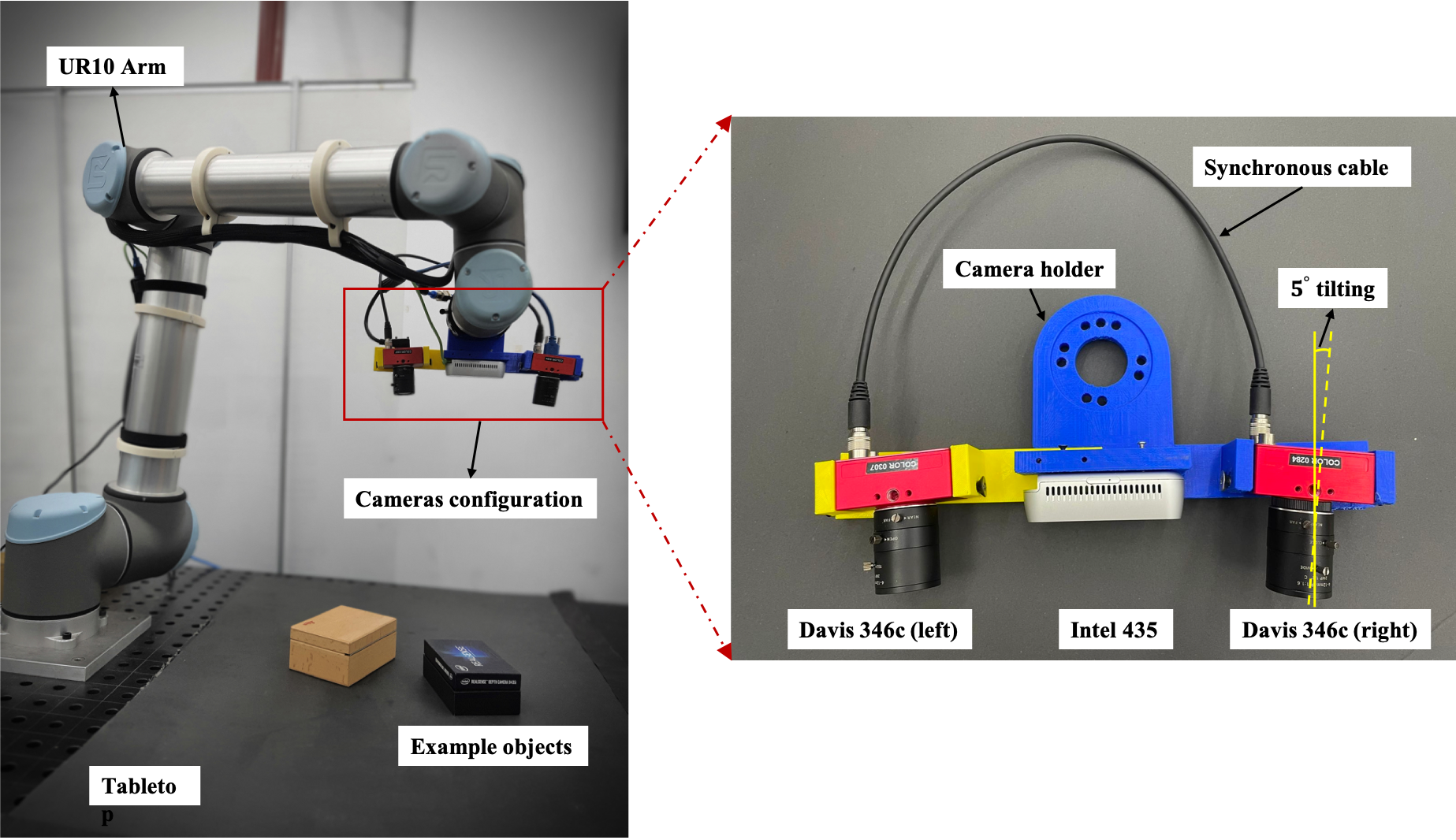}

\caption{Hardware setup. Experimental hardware setup (left-side figure): three cameras are fixed on the end-effector of the UR10's manipulator. 
Camera configuration (right-side figure): The RGBD camera \textit{Intel D435} is placed in the middle, and two event-based cameras \textit{Davis 346c} are mounted on the left and right sides with a tiled angle of 5 degrees towards the middle.}
\label{fig: dataset_setup}
\end{figure}

The hardware setup is built on the UR10 robot, as it can provide flexible and stable control of the camera's movement with positional repeatability of $0.1 \;mm$. Three cameras, including one RGBD camera \textit{Intel D435} and two event cameras \textit{Davis 346c}, are fixed in the camera holder 
attached to the robot's end-effector. 
The overview of the setup is illustrated in the left image of Figure \ref{fig: dataset_setup}.

To ensure the complete overlap of left and right cameras, the relative tilt angle between the two event cameras is 
calculated as 5 degrees with the assumed height of $0.82 m$. Therefore, the two event cameras are tiled with 5 degrees towards the RGBD camera. 
In addition, the two event-based cameras are synchronized by connecting synchronization connectors to ensure all of the events are triggered at the same rate. Due to the extremely high sample rate of a microsecond level, the synchronization between event cameras and the RGBD camera can be implemented by finding the nearest timestamps. 

\subsection*{Experimental protocol}
We collected ESD as two sub-sets: training (ESD-1) and testing (ESD-2) subsets for the unseen object segmentation task. Training and testing datasets consist of up to 10 objects and 5 objects respectively. Data sequences were collected under various experimental conditions which will be discussed in detail in subsection \textit{Dataset challenging factors and attributes}, including the different number of objects (2, 4, 6, 8, and 10 objects in ESD-1; 2 and 5 objects in ESD-2), lighting conditions (normal and low light), heights between cameras and tabletop (0.62\;m/s and 0.82\;m/s), occlusion conditions (with and without occlusion), cameras' moving speeds ($0.15\; m/s$, $0.3\; m/s$ and $1\; m/s$ in ESD-1, $0.15\; m/s$ and $1\; m/s$ in ESD-2) and trajectories (linear, linear + rotational and rotational in ESD-1; linear and rotational in ESD-2). Moreover, the objects in ESD-1 are different from ESD-2 and thus this dataset could be used in unknown object segmentation challenges. 

\begin{figure}[ht!]
\centering
\subfloat[linear movement]{\includegraphics[width = 2.2in]{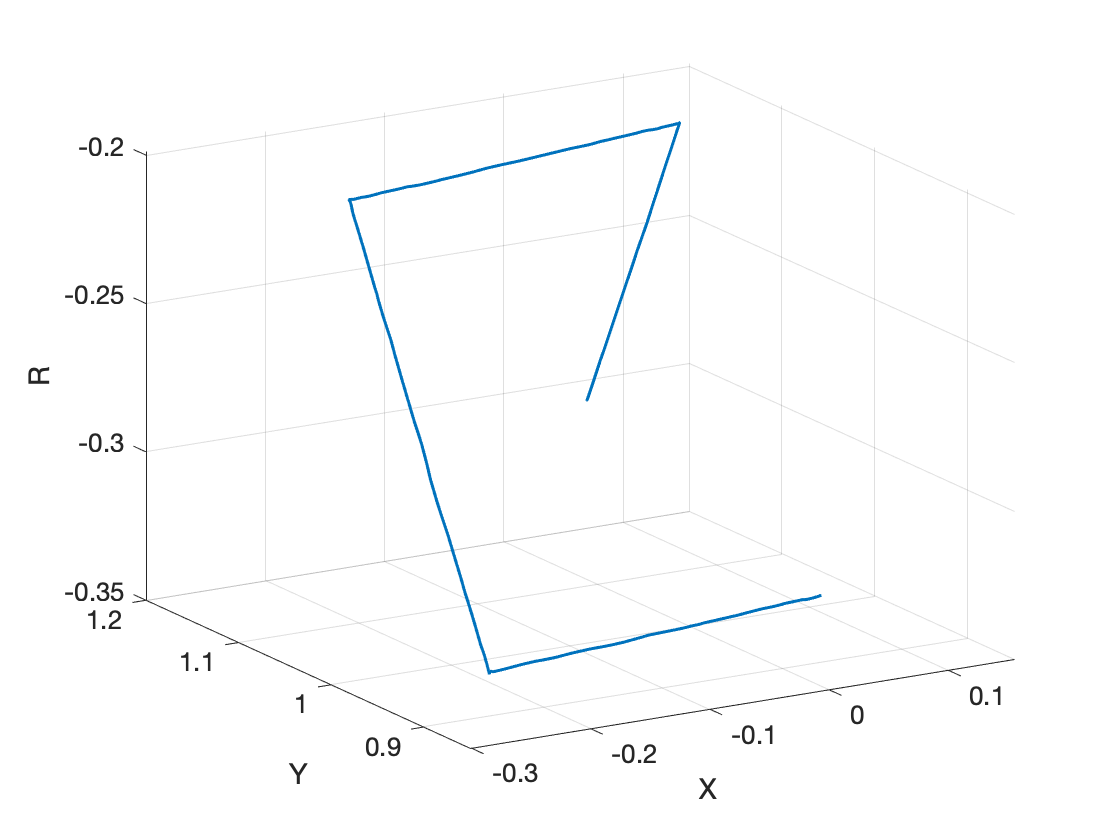}}
\subfloat[rotational movement]{\includegraphics[width = 2.2in]{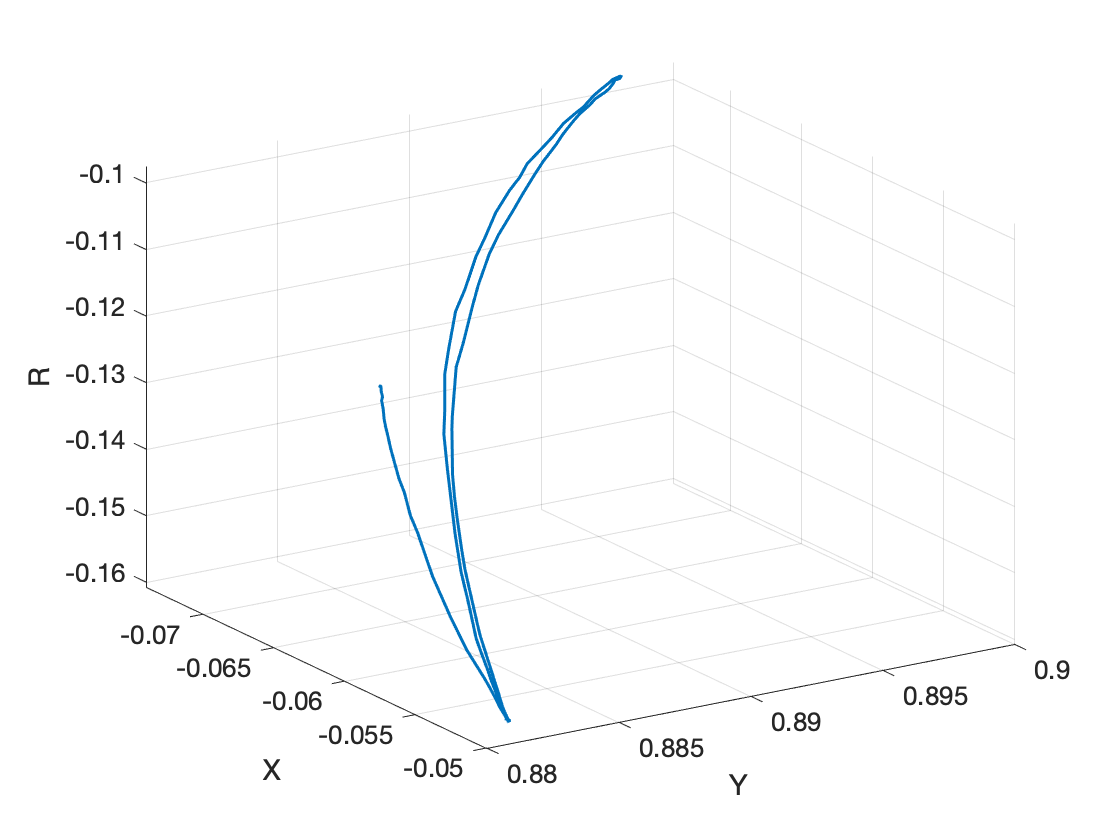}}
\subfloat[linear-rotational movement]{\includegraphics[width = 2.2in]{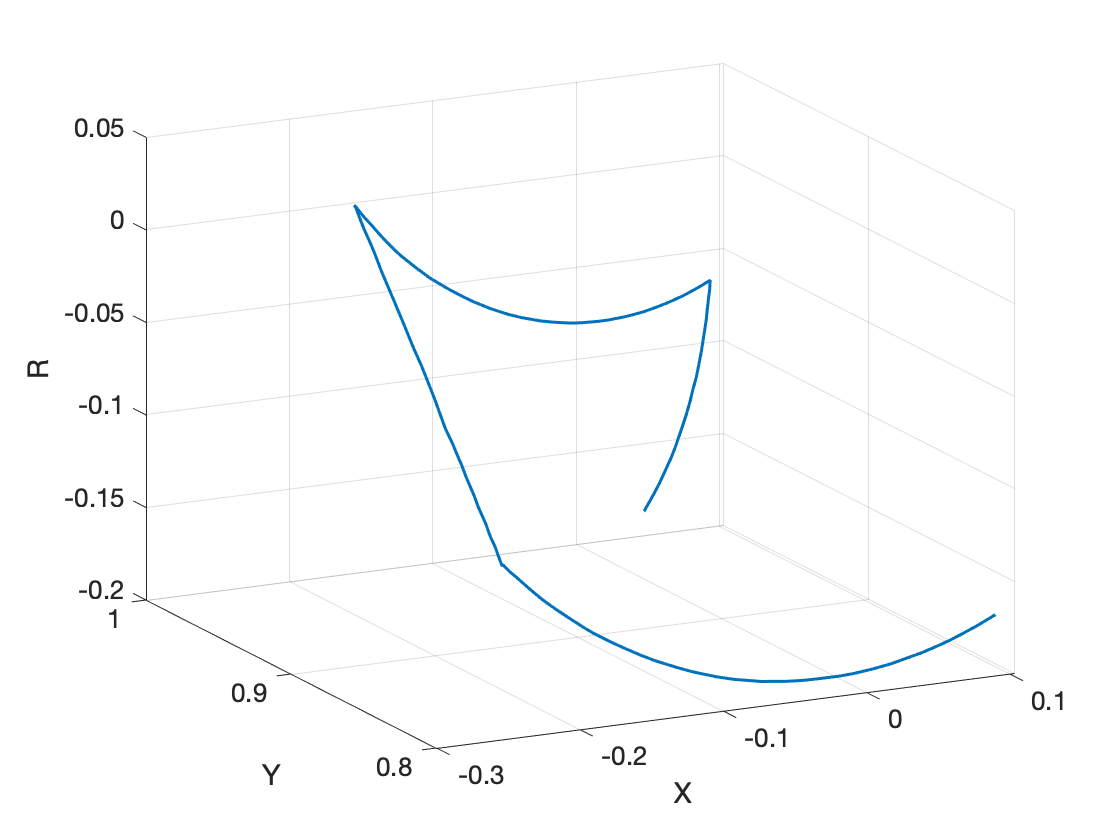}}
\caption{Designed moving trajectories in $x-y-r$ space, where $x-y$ indicates the plane that cameras move on, and the rotation is denoted in $r$ axis. }
\label{fig: movement}
\end{figure}

Before conducting experiments to collect data, all of the event cameras and the RGBD camera were calibrated to obtain the intrinsic and extrinsic parameters \cite{ayyad2023neuromorphic} for further data processing and annotation which will be elaborated in the following subsections. Then setting up the specific conditions for each particular experiment, such as the height of cameras and the lighting condition. Besides, the end-effector of UR10 robot that carried cameras moves to the setting point as starting position of all the experiments. Once all sets are ready, the cameras will move at various speeds and trajectories. Figure \ref{fig: movement} illustrates the three designed different moving trajectories in $x-y-r$ space using quaternion, where $x-y$ indicates the plane that the cameras move on, and the rotation is denoted in $r$ axis. 

Overall 115 experiments and 30 experiments were conducted for ESD-1 and ESD-2, respectively. The RGB part of the dataset consists of 14,166 annotated images. In total, 21.88 million and 20.80 million events from left and right event-based cameras are collected, respectively. 



\subsection*{Image and events annotation}

We tested different methods for the automatic annotation of RGB images and event data. Due to different features appearing with different perception angles of the camera, automatic labeling of RGB images with high accuracy is quite challenging. Thus, we manually labeled all RGB frames and reference them for event-based automatic annotation.  

\subsubsection*{Manual annotation of RGB frames}
Our proposed ESD dataset contains 11,361 images for training and 3,202 images for testing in total. 
We used the online web annotator CVAT \cite{cvat} to manually annotate the tabletop objects in each frame. 
CVAT offers automatic features for pixel labelling. The polylines tool is used to draw the boundaries around the objects. Dealing with occlusion is one of the challenges of annotating this dataset. The occluded object is declared as the background whereas the front object is declared as the foreground. 

\begin{figure}[ht!]
    \centering
    \includegraphics[scale=0.43]{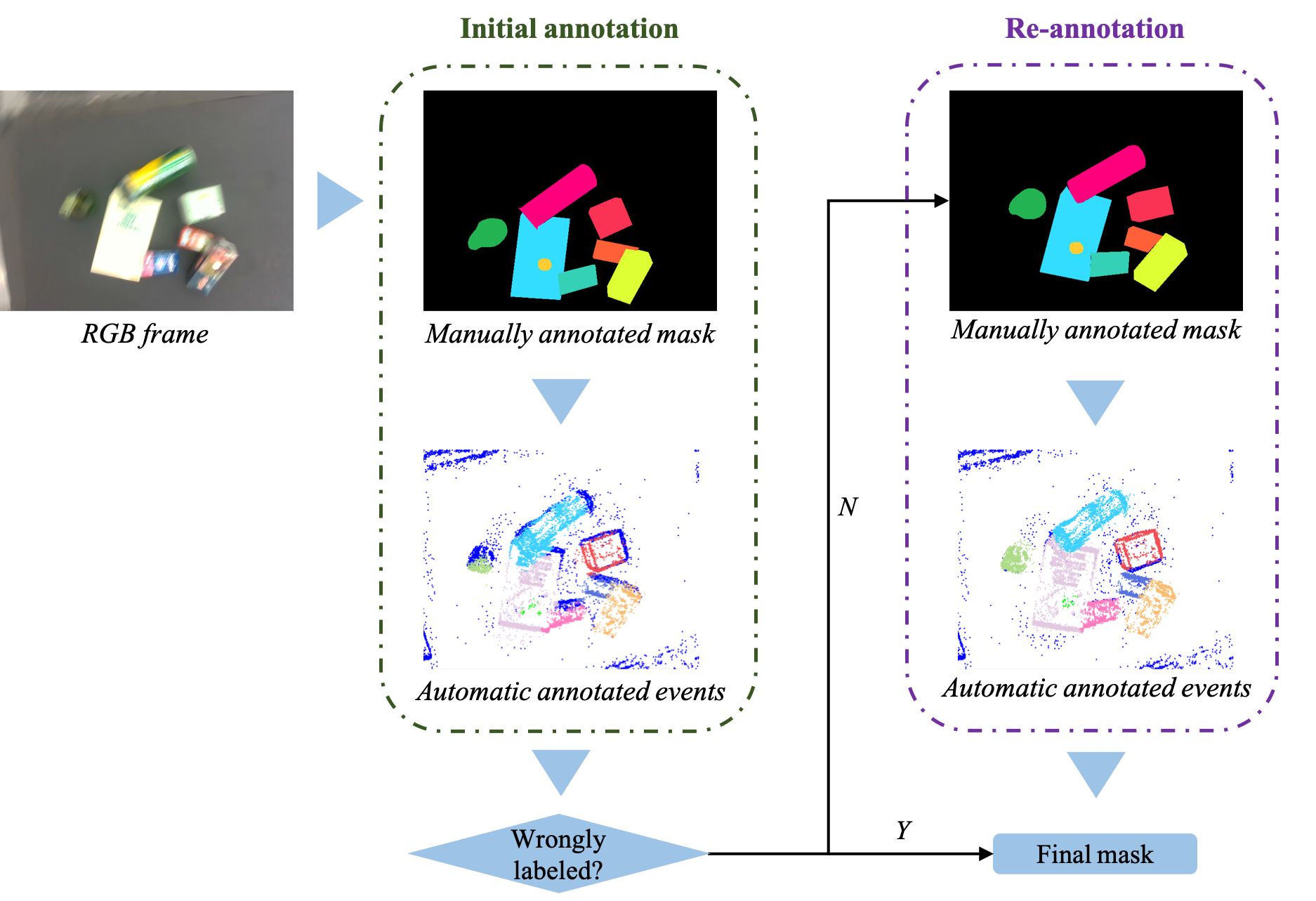}
    \caption{Two steps of labeling blurred images: initial annotation and re-annotation. 
    If wrong labels show in the event frame, the second-round labeling of the RGB mask will be triggered according to the initial annotated events. }
    \label{fig: events_annotation_steps}
\end{figure}

In addition, the motion blur of the RGBD camera caused by the low sampling rate introduces ambiguity for manual labeling of the objects' boundaries. Thus, we labeled blurred images in two steps as demonstrated in Figure \ref{fig: events_annotation_steps}; 
Initial annotation is based on the manually inferred objects' positions according to the cameras' moving trajectory. After the corresponding events are fitted and annotated as described in detail in section \textit{Automatic labeling of events data}, we can observe the events frame to understand whether the mask frame is well labeled with clear shapes and outlines of objects. If it shows precise annotation in the event frame, the initial annotated mask will be used as the final mask; otherwise, we will start the second step to re-label the RGB frame until the events are precisely annotated.

\subsubsection*{Automatic labeling of events data}
\label{sec: annotation_events}
Events are labeled according to the annotated RGB masks, 
and the Pseudo code of automatic annotation of a sequence of events captured in one experiment as described in Algorithm \ref{algo:auto_anno_events}. 

\begin{figure}[!ht]
\begin{algorithm}[H]

  \KwInput{Events stream: $E$ = (position $(x,y)$, polarity $p$, timestamp $ts_e$)\; 
  Manually labeled RGB frames: $S$\;
  Timestamps of RGB frames: $ts$\;
  Depth information from RGBD camera: $d$\;
  Number of events per Iterative Closest Point (ICP) \cite{besl1992method} process: $n$\;
  Total number of frames in the sequence $N$ \;
  The intrinsic and extrinsic parameters of cameras \;}
  
  \KwOutput{Annotated events data with depth $d_e$ and label $l_e$ as $E' = (x, y, p, ts_e, d_e, l_e)$}
  Initialize the iteration index $i=0$\;
\While{$i \leq N$}{
  Find start index $I_0$ of closest timestamp between $ts_e$ and $ts_i$\;
  Find end index $I_1$ of closest timestamp between $ts_e$ and $ts_{i+1}$\;
  Events between two frames: $E_i = E(I_0:I_1)$\;
  Obtain edge image $K_i$ of RGB frame by Canny Edge detection\;
  Transform edge image $K_i$ into event-based camera coordinate as $K_{ei}$ (Equation \eqref{eqn: p_to_c}-\eqref{eqn: c_to_p})\;
  Transform mask $S_i$ into event-based camera coordinate as $S_{ei}$ (Equation \eqref{eqn: p_to_c}-\eqref{eqn: c_to_p})\;
  Dilate $S_{ei}$\;
  \For{\texttt{<j=1:n:the number of events>}}
  {
    Perform ICP to fit the transformed event mask $S_{ei}$ into events in sub-interval $E_{ij}$ and find the rigid transformation $\mathbf{R}$, $\mathbf{T}$;\\
    Transform $S_{ei}$ into events in sub-interval $S_{eij} = [R \; T]S_{ei}$;\\
    Inherit the depth and label of each event, according to the transformed manually labeled mask images $S_{eij}$;\\
    Automatically label the events as $E'_{ij}$;\\
    Store $E'_{ij}$ into $E'$;
  }
  $i=i+1$
  }
\caption{Automatic Annotation for Events Data}
\label{algo:auto_anno_events}
\end{algorithm}
\end{figure}

Events recorded can be considered as a continuous data stream with a high frequency (few microseconds). Thus, we divided sequences of events into intervals "$E$" of around $60\; ms$ which is the same sampling period of the RGBD camera by finding the nearest timestamp between events and the RGB frame. 
Simultaneously, annotated mask frames "$\textit{S}$" in RGBD coordinate are transformed to events coordinates as "$\textit{S}_e$" as described in Equation \eqref{eqn: p_to_c}-\eqref{eqn: c_to_p}. First, the forward projection is applied to transform mask frames "\textit{S}" in RGBD coordinate into RGBD camera coordinate as "$\textit{S}_c$" using the camera intrinsic parameters (Equation \eqref{eqn: p_to_c}). As expressed in Equation \eqref{eqn: c_to_c}, the coordinate transformation is applied twice to transform "$\textit{S}_c$" into world coordinate and event camera coordinate in sequence using the cameras' extrinsic parameters. Building on that, masks in event camera coordinate are backward projected into events coordinate as described in Equation \eqref{eqn: c_to_p}. 

\begin{equation}
    \begin{cases}
    x= (u-c_x)z/f_x\\
        y = (v-c_y)z/f_y\\
    \end{cases}
\label{eqn: p_to_c}
\end{equation}
\begin{equation}
    \begin{bmatrix}
        x_e\\
        y_e\\
        z_e\\
        1
    \end{bmatrix} = \begin{bmatrix}
        \mathbf{R_e} & \mathbf{T_e}\\
        0 & 1 
    \end{bmatrix}^{-1}
    \begin{bmatrix}
        \mathbf{R} & \mathbf{T}\\
        0 & 1 
    \end{bmatrix}
    \begin{bmatrix}
        x\\
        y\\
        z\\
        1
    \end{bmatrix}
\label{eqn: c_to_c}
\end{equation}
\begin{equation}
    \begin{cases}
    u_e= f_{xe}x_e/z_e+c_{xe}\\
        v_e = f_{ye}y_e/z_e+c_{ye}\\
    \end{cases}
\label{eqn: c_to_p}
\end{equation}
where $(x, y, z)$, $(u, v)$, $(x_e, y_e, z_e)$, $(u_e, v_e)$ and $(X, Y, Z)$ represent the same point in RGBD camera coordinate, RGB image plane, event camera coordinate, events image plane, and world coordinate systems respectively. ${c_x, c_y}$ and ${c_{xe}, c_{ye}}$, indicate the center points in RGB and events image planes, respectively. Similarly, ${f_x, f_y}$ and ${f_{xe}, f_{ye}}$ denote the focal length of RGBD and event camera, respectively. $\mathbf{R}$ and $\mathbf{T}$ express the rotation matrix and translation vector from the RGBD camera coordinate to the world coordinate system. $\mathbf{R_e}$, $\mathbf{T_e}$ describes the rotation matrix and translation vector from the event camera coordinate to the world coordinate system. 

However, the events recorded asynchronously appear in different locations  
since the camera keeps moving. Thus, 
events between two consecutive RGB frames are sliced into sub-intervals with 300 events "$E_{ij}$". Then fitting transformed event mask "$S_e$" into events coordinate as "$S_{et}$"  by applying the Iterative Closest Point (ICP) algorithm \cite{besl1992method} to find the rigid body transformation between the two corresponding point sets $X=\{x_1, x_2, .., x_n\}$ and $P=\{p_1, p_2, .., p_n\}$. The ICP algorithm assumes that the corresponding points $x_i$ and $p_i$ are the nearest ones, so the working principle is to find the rotation matrix $R$ and translation $t$ that minimizes the sum of the squared error $E(\mathbf{R}, \mathbf{T})$ as expressed in Equation \eqref{eqn: ICP}.  
\begin{equation}
    E(R,T) = \frac{1}{N_p}\sum_{i=1}^{N_p}\Vert x_i - R_{p_i} - t\Vert^2 
\label{eqn: ICP}
\end{equation}

\begin{figure}[h!]
    \centering
    \includegraphics[scale=0.34]{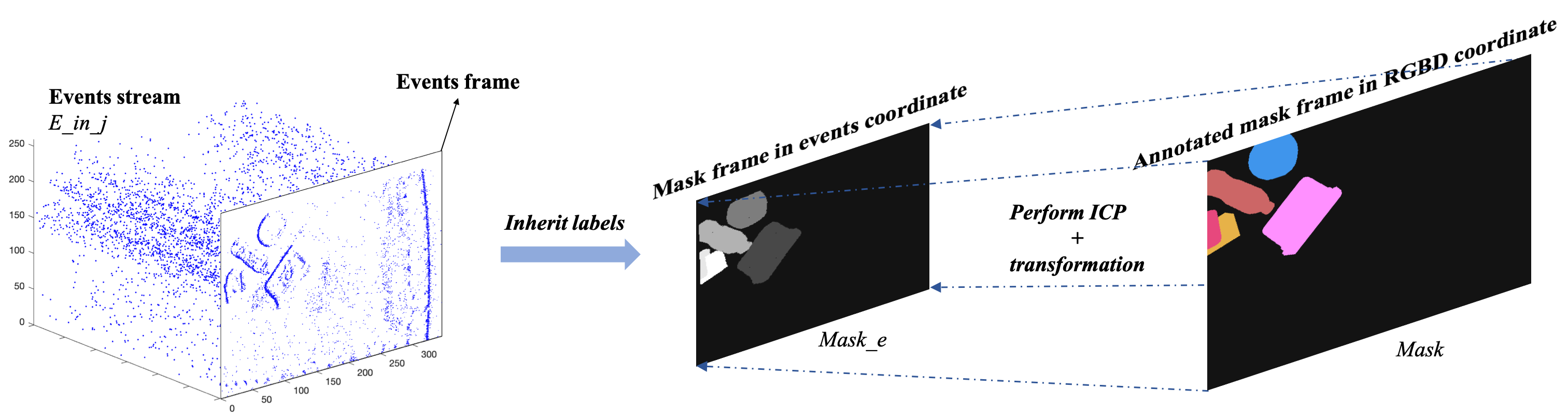}
    \caption{Principle of mapping the interval of events on the RGB frame coordinate for annotation}
    \label{fig: ICP}
\end{figure}

According to the transformation matrix of rotation and translation calculated using ICP, the corresponding location of events on annotated mask frames will be obtained as $x'_i=[\mathbf{R} \; \mathbf{T}]x_i$.
Therefore, the label of pixels on the RGB mask will be inherited to events. The working principle is illustrated in Figure \ref{fig: ICP}.   

However, two events cameras and an RGBD camera are placed in different positions in our setup as shown in Figure 1, their views might not be fully overlapped due to the limitation of the distance between objects and the camera. Thus, the area only sensed by the event camera will be labeled as background, even if it corresponds to an object. To tackle this issue, we crop out the blind area which is only sensed by the event camera.

\subsection*{Data Visualization}
\begin{figure*}[ht!]
\centering
\rotatebox{90}{2 objects}
\subfloat{\includegraphics[width = 1.45in,height=1.3in]{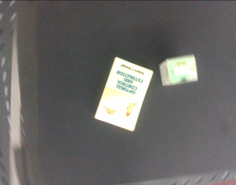}} 
\subfloat{\includegraphics[width = 1.5in,height=1.3in]{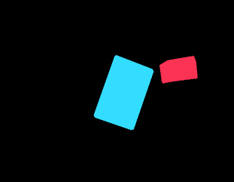}}
\subfloat{\includegraphics[width = 1.5in,height=1.3in]{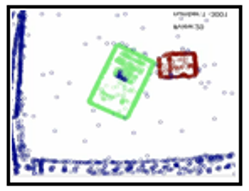}} \\
\rotatebox{90}{4 objects}
\subfloat{\includegraphics[width = 1.5in,height=1.3in]{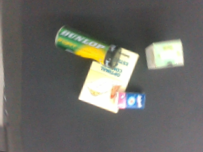}}
\subfloat{\includegraphics[width = 1.5in,height=1.3in]{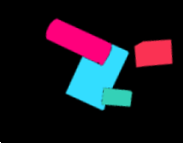}}
\subfloat{\includegraphics[width = 1.5in,height=1.3in]{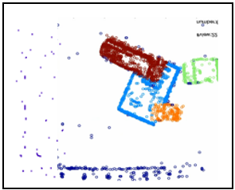}}\\
\rotatebox{90}{6 objects}
\subfloat{\includegraphics[width = 1.5in,height=1.3in]{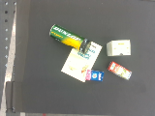}}
\subfloat{\includegraphics[width = 1.5in,height=1.3in]{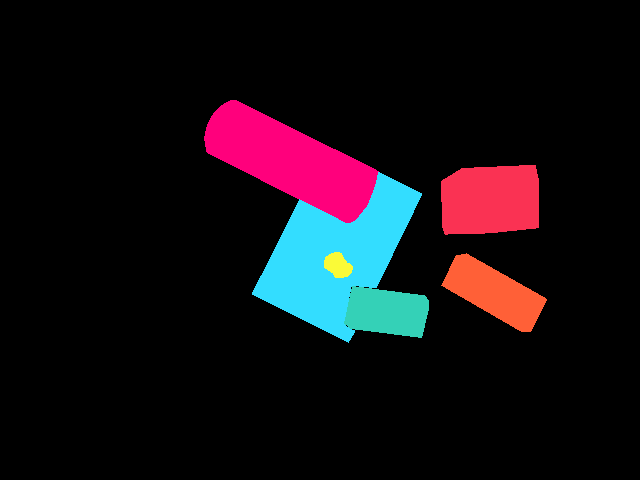}}
\subfloat{\includegraphics[width = 1.5in,height=1.3in]{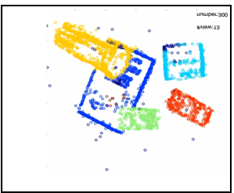}}\\
\rotatebox{90}{8 objects}
\subfloat{\includegraphics[width = 1.5in,height=1.3in]{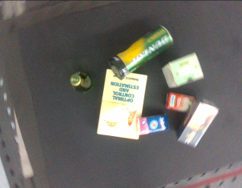}}
\subfloat{\includegraphics[width = 1.5in,height=1.3in]{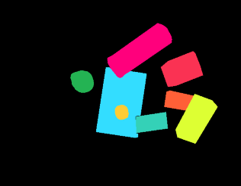}}
\subfloat{\includegraphics[width = 1.5in,height=1.3in]{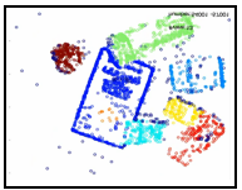}}\\
\rotatebox{90}{10 objects}
\subfloat[raw RGB images]{\includegraphics[width = 1.5in,height=1.3in]{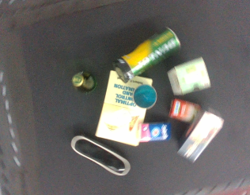}} 
\subfloat[Ground truth]{\includegraphics[width = 1.5in,height=1.3in]{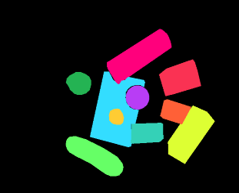}}
\subfloat[annotated events]{\includegraphics[width = 1.5in,height=1.3in]{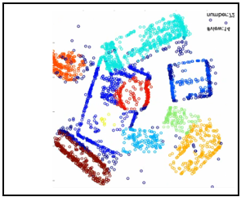}}

\caption{Example of the ESD-1 in terms of the number of objects attributes, under the condition of 0.15 moving speed, normal light condition, linear movement, and 0.82 height. Different colors in the RGB ground truth and annotated event mask mean different labels. Better view in color.}
\label{fig: data_visualization_ESD1}
\end{figure*}

\begin{figure*}[h!]
\centering
\rotatebox{90}{2 objects}
\subfloat{\includegraphics[width = 1.5in,height=1.3in]{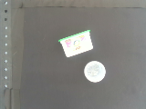}} 
\subfloat{\includegraphics[width = 1.5in,height=1.3in]{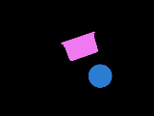}} 
\subfloat{\includegraphics[width = 1.5in,height=1.3in]{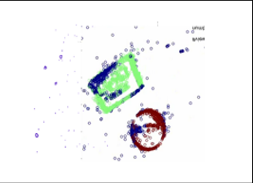}} \\
\rotatebox{90}{5 objects}
\subfloat[raw RGB images]{\includegraphics[width = 1.5in,height=1.3in]{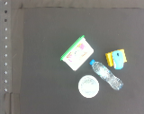}}
\subfloat[Ground truth]{\includegraphics[width = 1.5in,height=1.3in]{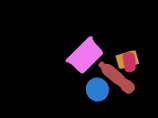}}
\subfloat[annotated events]{\includegraphics[width = 1.5in,height=1.3in]{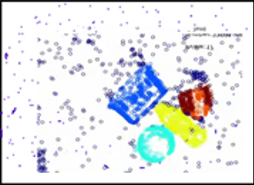}}\\
\caption{Example of unknown objects ESD-2 dataset in terms of the number of objects attributes, under the condition of 0.15 moving speed, normal light condition, linear movement, and 0.82 height. Different colors in the RGB ground truth and annotated event mask mean different labels. Better viewed in color.}
\label{fig: data_visualization_ESD2}
\end{figure*}

We divided ESD into training (ESD-1) and testing (ESD-2) subsets for unseen object segmentation tasks. Training and testing dataset consists of up to 10 objects and 5 objects respectively. The testing dataset also challenges solving unseen object segmentation tasks since it contains different objects to ESD-1 the training dataset. Data sequences were collected under various experimental conditions which will be discussed in detail in subsection \textit{Dataset challenging factors and attributes}. 
Examples of ESD-1 in terms of the number of objects' attributes can be visualized in Figure \ref{fig: data_visualization_ESD1}. The raw RGB image, annotated mask, and the corresponding annotated events ($N=3000$) are illustrated for conditions of the different number of objects. Particularly, in the clusters of 2 objects, both the objects (ie. Book and box) are distanced from each other. For clusters of more than 2 objects, there are occlusions among objects. Besides, examples of ESD-2 for unseen object segmentation in terms of other attributes are depicted in Figure \ref{fig: data_visualization_ESD2}. 

\section*{Data Records}
\begin{figure}[ht!]
    \centering
    \includegraphics[scale=0.54]{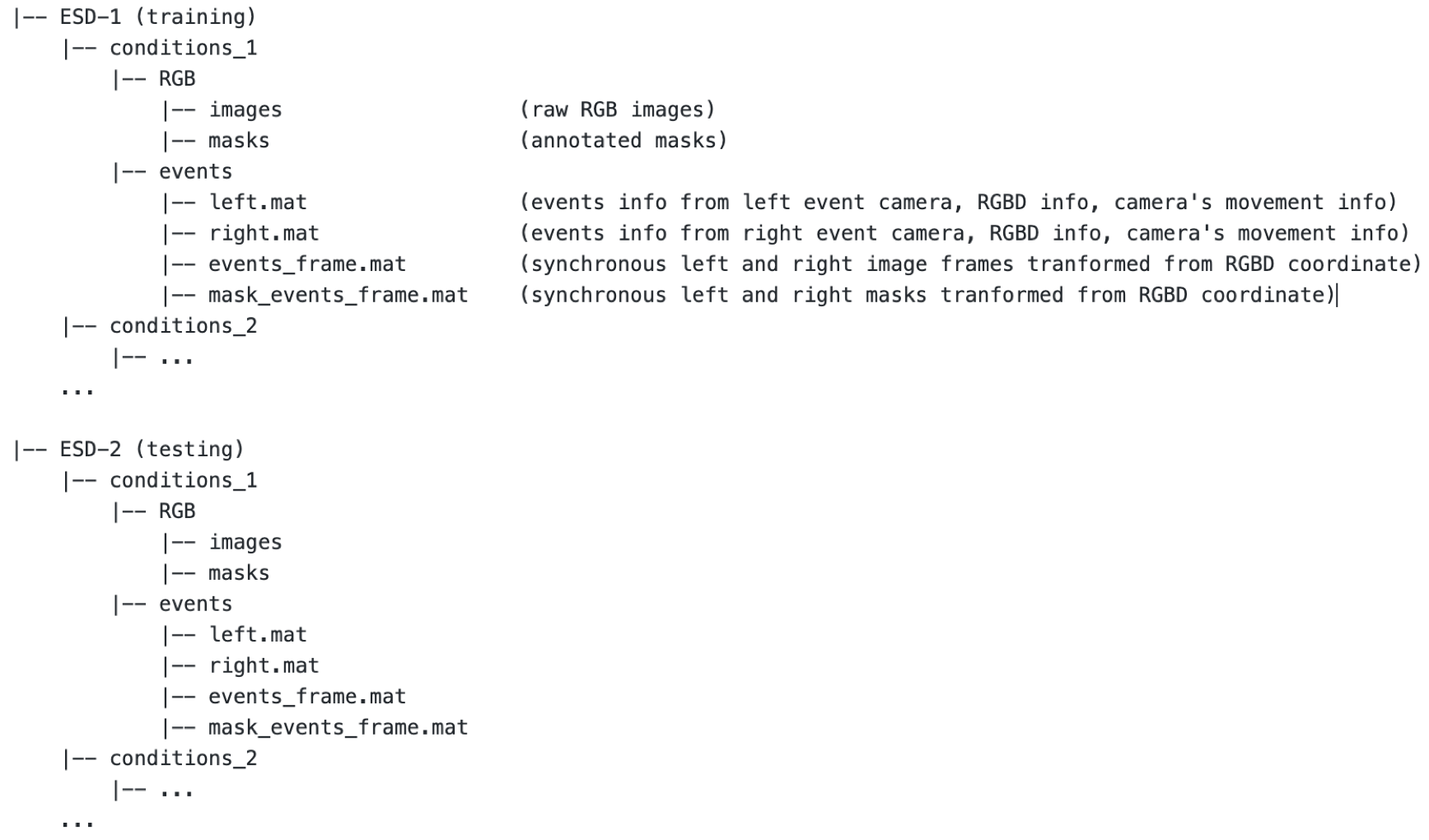}
    \caption{Dataset structure. Each sequence was recorded in the "events" subfolder under different experimental conditions with a unique name under the training or testing path. Event-related and  frame-related information is stored under "events" and "RGB" folders, respectively. Particularly, raw images and annotated masks are contained in the "RGB" subfolder under different experimental conditions. Events with RGBD information of both event cameras, image, and mask frames converted from RGBD coordinates and cameras' movement are recorded under the "events" folder.
    }
    \label{fig: structure}
\end{figure}

All of the data is available at Open Science Framework at Figshare \cite{ESD} including ESD-1 \cite{ESD-1} and ESD-2 \cite{ESD-2}, whose structure is demonstrated in Figure \ref{fig: structure}. 
\subsection*{Data format}
Event-related data in the same sequence is recorded in four files under the "events" folder of certain conditions. Events information from left \textit{Davis 346c}, RGBD information from \textit{Intel 435}, and information of cameras' movement is saved in \textit{left.mat}. Similarly, \textit{right.mat} contains events and frames information from the right event-based camera and RGBD camera and information of cameras' movement. Additionally, synchronous image frames and mask frames converted from RGBD camera coordinates for both event-based camera coordinates are reserved in \textit{events\_frame.mat} and \textit{mask\_events\_frame.mat}. Moreover, the raw RGB images and ground truth masks are also provided in the "RGB" folder of all experimental conditions.  

\subsection*{Dataset challenging factors and attributes}\label{sec: attributes}
We constructed ESD dataset with various scenarios and challenges in the indoor cluttered environment. 
We briefly define the attributes as below, and the symbol \textbf{*} is varying in specific conditions:
\begin{itemize}
     \item \textbf{Various number of objects (O*):} The complexity of the scene can be affected by the number of objects. Thus, we selected different numbers of objects with various shapes and layouts to increase the diversity of the scenes. Particularly, scenes of 2, 4, 6, 8, and 10 objects are collected in ESD-1. Scenes of 2 and 5 objects are collected in ESD-2. 
     \item \textbf{Cameras' moving speed  (S*):} Motion blur is an open challenge in computer vision tasks. We collected data with different moving speeds (S015: $0.15 m/s$, S03: $0.3 m/s$, S1: $1 m/s$) of cameras to introduce various degrees of motion blur of RGB frames. 
     
     \item \textbf{Cameras' moving trajectory (M*): } From the observation, events may not be captured if they are on the edge which is parallel to the camera's moving direction that is challenging in event-based processing. We introduce this attribute as linear (ML), rotational (MR), and linear-rotational ((MLR)) moving trajectory to cover all the edge directions.  
     \item \textbf{Illumination Variant (*L):} Illumination has a substantial impact on an object’s appearance and is still an open challenging problem in segmentation. In the dataset recording, we collected data in low lighting (LL) and normal lighting (NL) conditions. 
     \item \textbf{Height between tabletop and cameras (*H):} It affects the size of the overlap area of stereo cameras. Thus, we introduce it as one of the attributes. The dataset is collected with a higher height (HH) and a lower height (LH) indicating that the sensing areas from the stereo camera are fully overlapped and partially overlapped, respectively.
     \item \textbf{Occlusion condition (*O): } Occlusion is a classical and challenging scenario in segmentation, that is caused by the integration of objects in the scene. We placed objects with occlusion (OY) and without occlusion (ON). 
\end{itemize}

\begin{figure*}[h!]
\centering
\rotatebox{90}{Low lighting condition}
\subfloat{\includegraphics[width = 1.5in,height=1.5in]{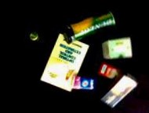}} 
\subfloat{\includegraphics[width = 1.5in,height=1.5in]{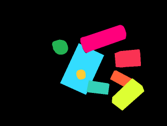}}
\subfloat{\includegraphics[width = 1.5in,height=1.5in]{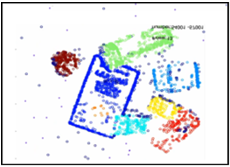}} \\
\rotatebox{90}{Fast camera motion}
\subfloat{\includegraphics[width = 1.5in,height=1.5in]{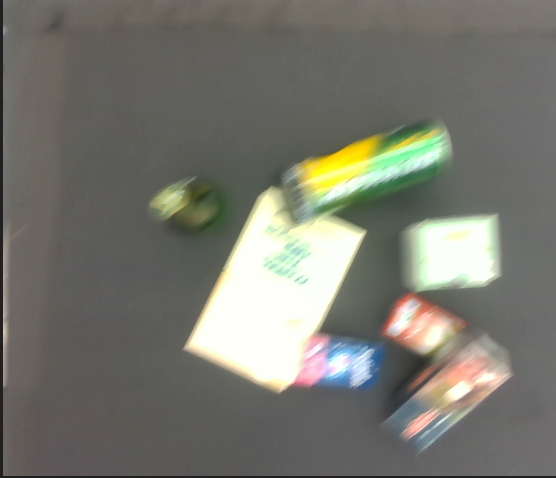}}
\subfloat{\includegraphics[width = 1.5in,height=1.5in]{figures/Images_degradation_RGB_Mask_82.png}}
\subfloat{\includegraphics[width = 1.5in,height=1.5in]{figures/Images_degradation_Event_Mask_82.png}}\\
\rotatebox{90}{Occlusion}
\subfloat{\includegraphics[width = 1.5in,height=1.5in]{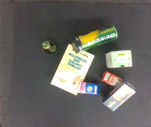}}
\subfloat{\includegraphics[width = 1.5in,height=1.5in]{figures/Images_degradation_RGB_Mask_82.png}}
\subfloat{\includegraphics[width = 1.5in,height=1.5in]{figures/Images_degradation_Event_Mask_82.png}}\\
\rotatebox{90}{Lower height}
\subfloat[raw RGB images]{\includegraphics[width = 1.5in,height=1.5in]{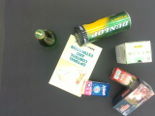}} 
\subfloat[Ground truth]{\includegraphics[width = 1.5in,height=1.5in]{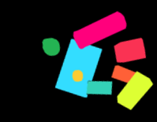}}
\subfloat[Annotated events
]{\includegraphics[width = 1.5in,height=1.5in]{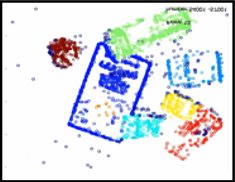}}
\caption{Sample images of RGB, masks and annotated events are selected from our proposed ESD dataset. (a) Shows tabletop objects under low lighting conditions. (b) Shows the motion blur scenarios because of the fast camera motion with 1m/s speed. (c) Shows the objects are occluded by others. (d) Shows the lower height of cameras with 0.62m from the tabletop. Different colors in the RGB ground truth and
annotated event masks mean different labels. Better view in color.}
\label{fig: data_visualization_2}
\end{figure*}

\begin{figure}[ht!]
\centering
\subfloat[Sequences statistic]{\includegraphics[width = 3.2in]{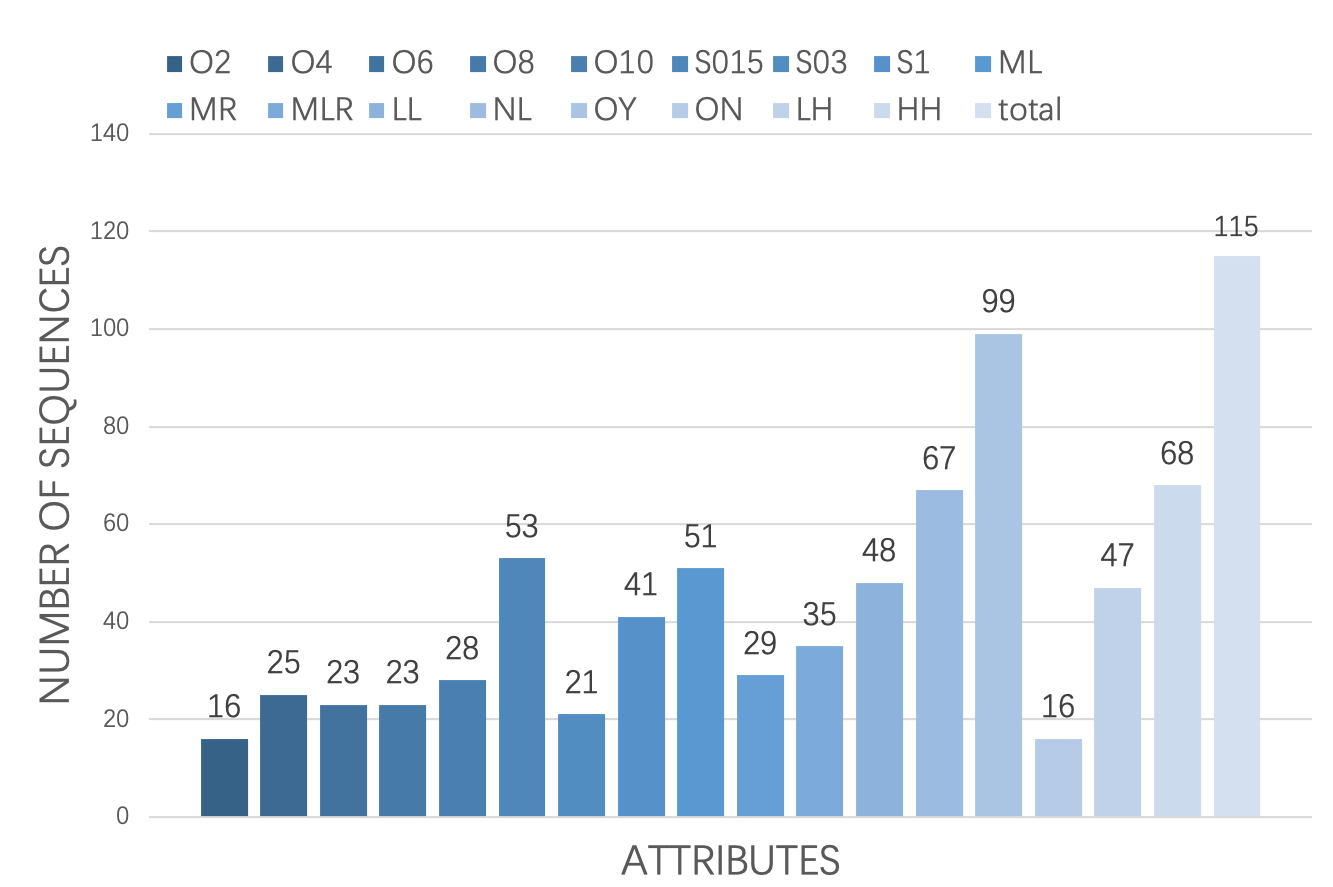}}
\subfloat[Frames statistic]{\includegraphics[width = 3.2in]{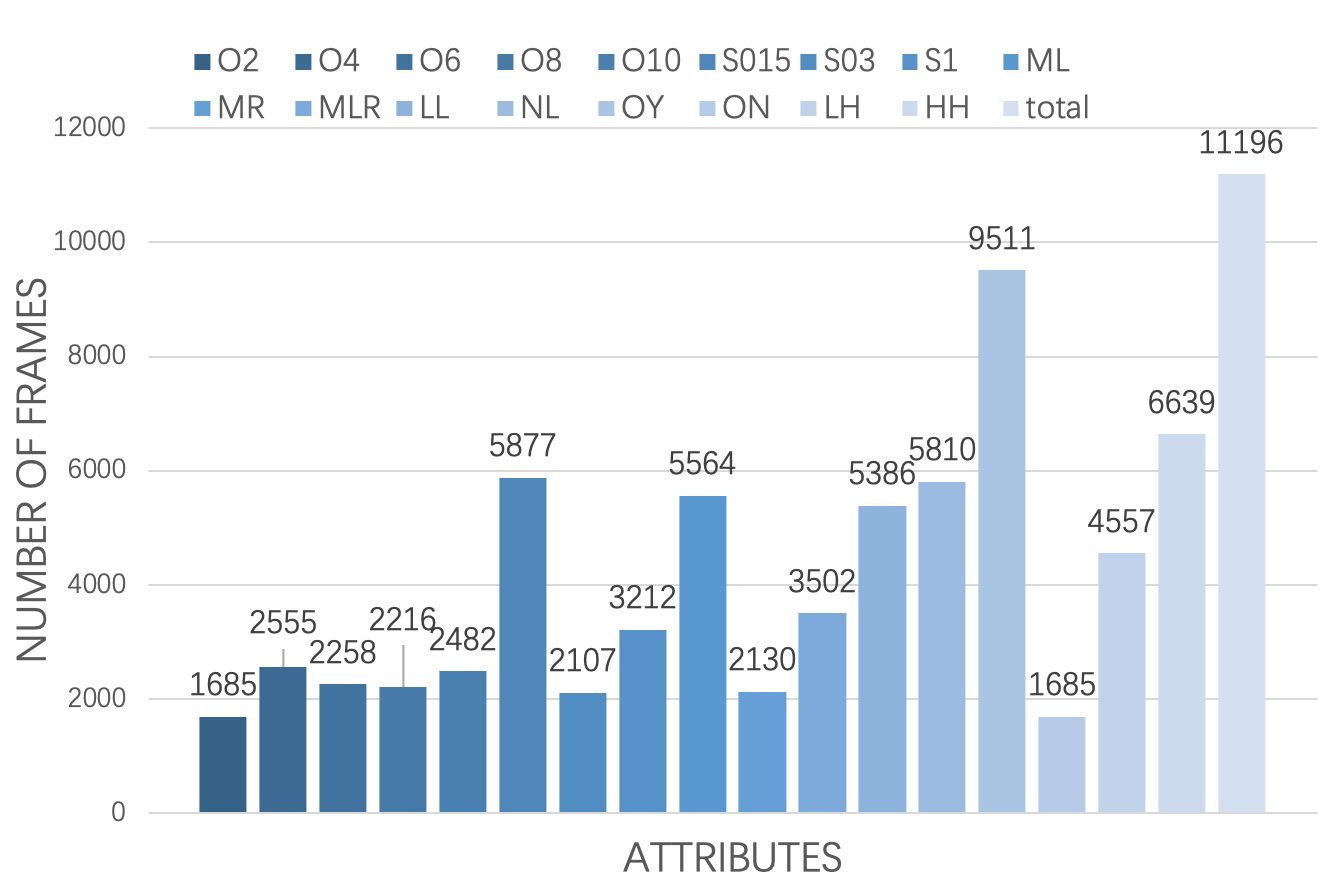}}\\
\subfloat[Events statistic]{\includegraphics[width = 6.4in]{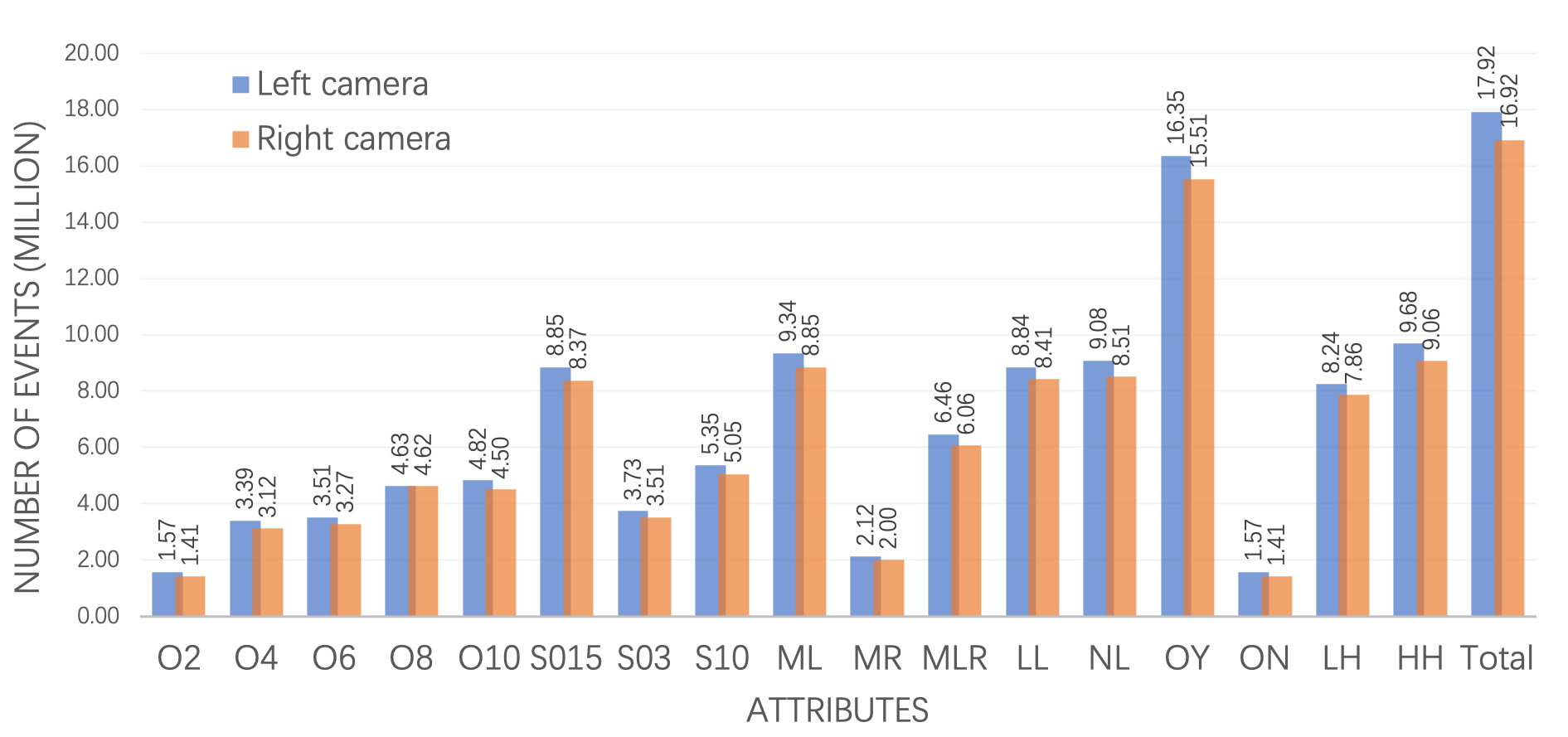}}
\caption{
ESD-1 statistic: sequence (a), frames (b) and events (c) statistic in terms of attributes. ML, MR and MLR indicate linear, rotation and hybrid moving types; LN and LL represent normal and low light conditions; S015, S03, and S1 describe the camera’s moving speed of 0.15 m/s, 0.3 m/s and 1 m/s; Similarly, O2, O4, O6, O8 and O10 express sequences of 2-10 objects; The occlusion cases are with and without occlusion referred as OY and ON, respectively. Additionally, the total quantities of sequences, frames and events are also presented in (a)-(c), respectively. Better viewed in color. }
\label{fig: sequence statistic_1}
\end{figure}

\begin{figure}[ht!]
\centering
\subfloat[Sequences statistic]{\includegraphics[width = 3.2in]{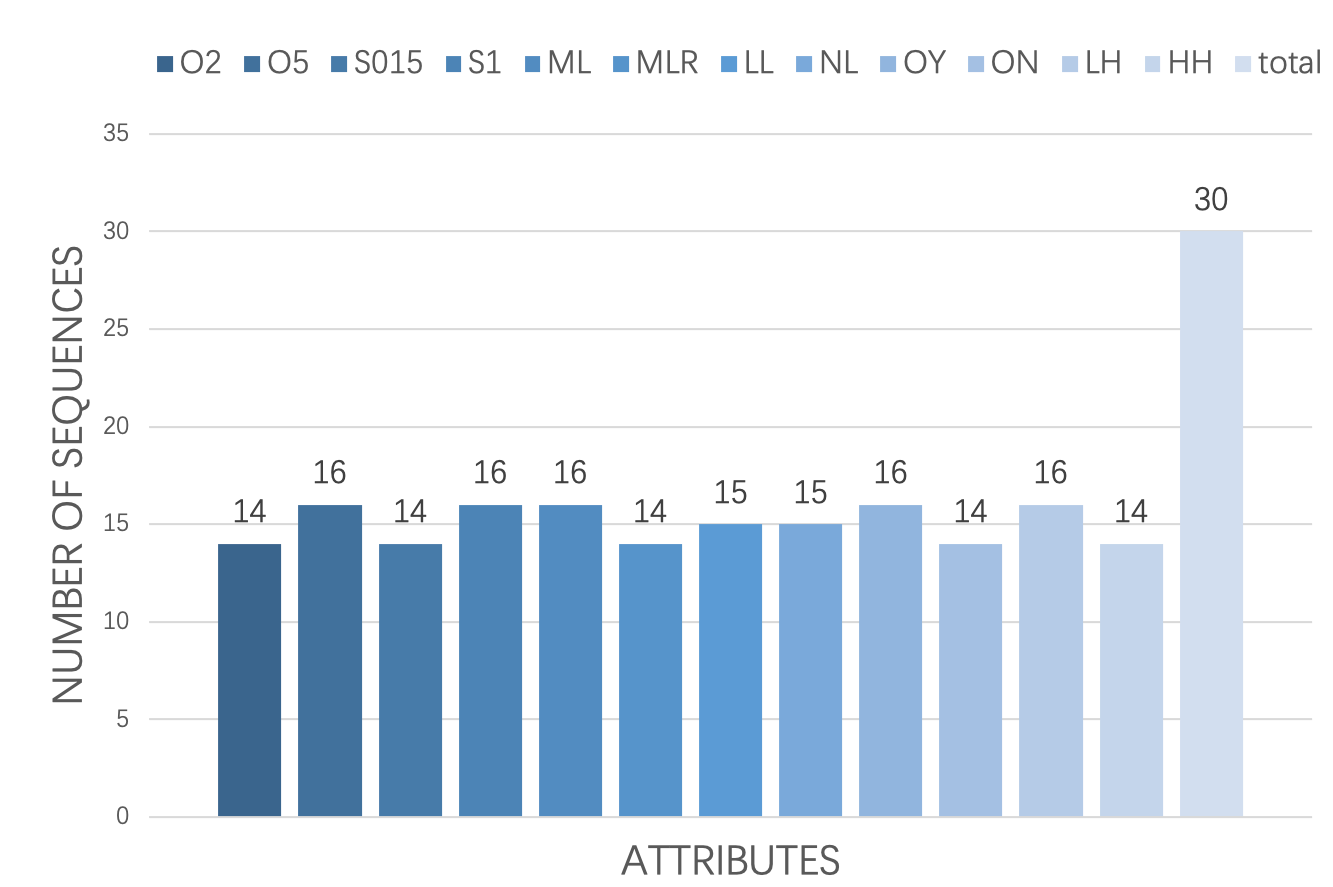}}
\subfloat[Frames statistic]{\includegraphics[width = 3.2in]{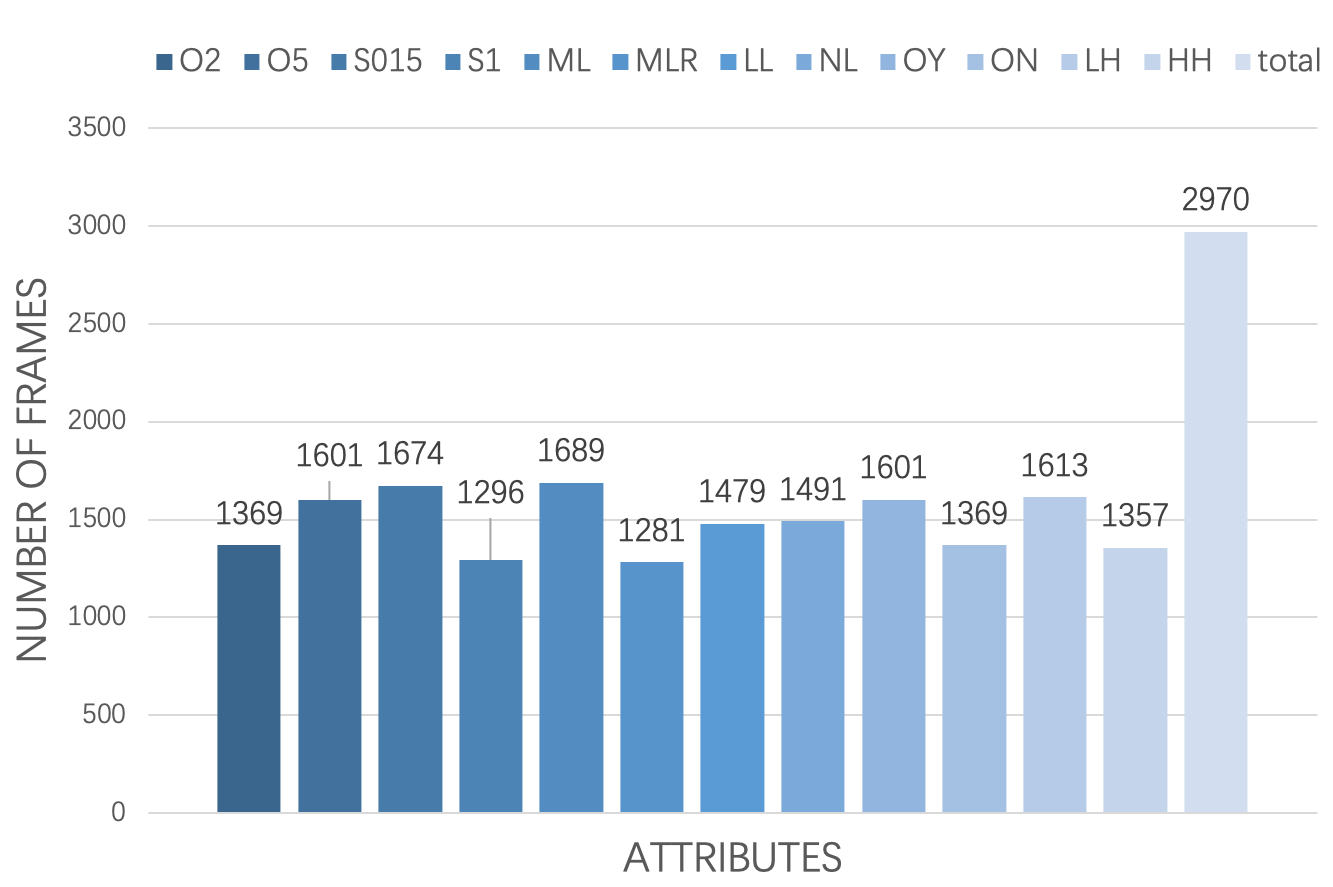}}\\
\subfloat[Events statistic]{\includegraphics[width = 6.4in]{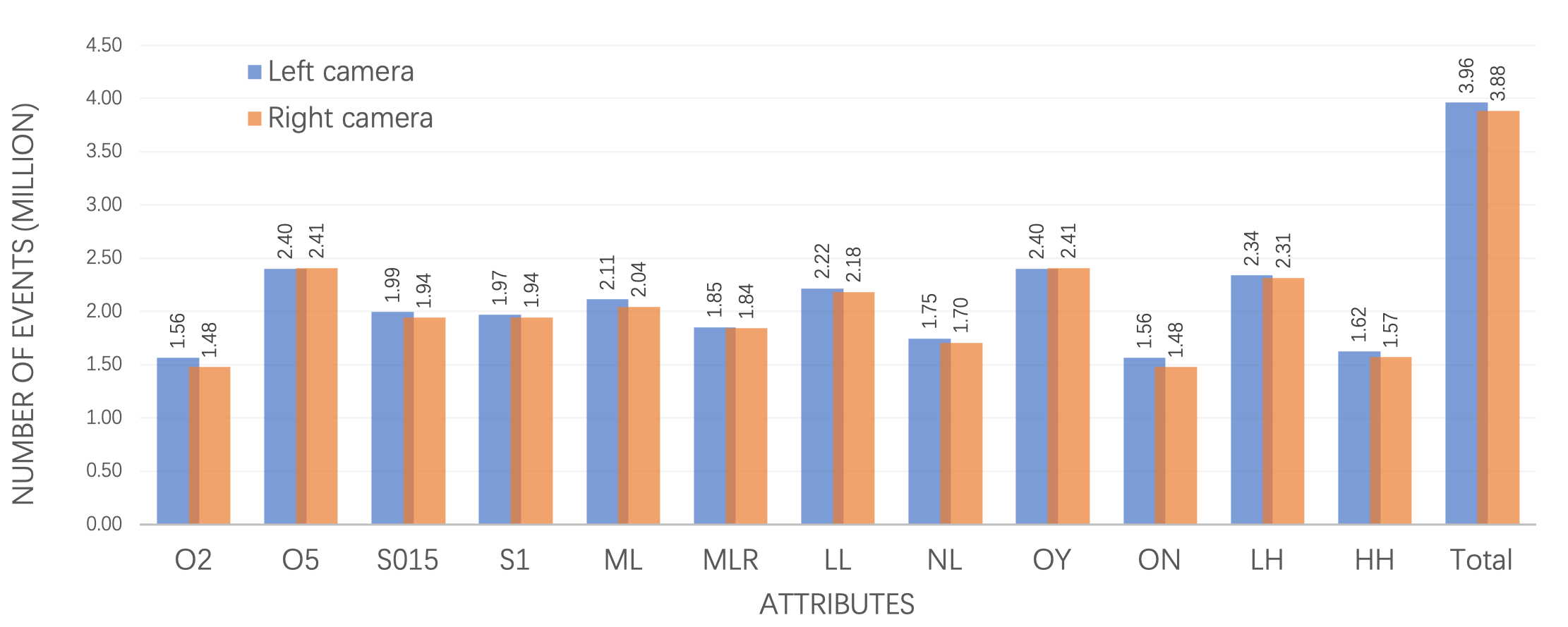}}
\caption{
ESD-2 statistic: sequence (a), frames (b) and events (c) statistic in terms of attributes. ML and MLR indicate linear and hybrid moving types; LN and LL represent normal and low light conditions; S015 and S1 describe the camera’s moving speed of 0.15 m/s and 1 m/s; Similarly, O2 and O5 express sequences of 2 and 5 objects; The occlusion cases are with and without occlusion referred as OY and ON, respectively. Additionally, the total quantities of sequences, frames and events are also presented in (a)-(c), respectively. Better view in color. }
\label{fig: sequence statistic_2}
\end{figure}

Sample images taken from our proposed ESD dataset with various attributes are shown in Figure \ref{fig: data_visualization_2}.
To meet the requirements of applications for unknown objects, two subsets are collected namely ESD-1 and ESD-2 with different objects. Thus, ESD-2 can be utilized as unknown objects dataset to test the performance. A total of 115 sequences are collected and labeled in ESD-1, and their attributes are statisticized in Figure \ref{fig: sequence statistic_1} including different light conditions, moving speed, moving trajectories and objects with occlusion. Similarly, 30 sequences are collected and labeled in ESD-2, and the data statistics are depicted in Figure \ref{fig: sequence statistic_2}. Eight topics are involved in each sequence of data: end effector’s pose and moving velocity, RGB frames and depth maps from D435, RGB frames and events stream from left and right Davis 346C. 

\section*{Technical Validation}

\subsection*{Evaluation metrics}
Our dataset ESD provides labels of events for individual objects that can be used in instance segmentation tasks. Moreover, objects in ESD are of different categories, so they can also be utilized in semantic segmentation. 
In this work, we evaluate our dataset using instance/semantic segmentation approaches. Therefore, the standard metrics for 
segmentation: accuracy and mean Intersection over Union (mIoU) are utilized to quantify testing results. 
Pixel accuracy is a metric to calculate the percent of pixels classified correctly as expressed in Equation \eqref{eqn: acc}. 

\begin{equation}
    Acc(p,p\prime) = \frac{1}{N}\sum_i^N\delta(p_i,p\prime_i)
    \label{eqn: acc}
\end{equation}

\noindent where $p$, $p\prime$, $N$, and $\delta$ represent the ground truth image, the predicted image, the total number of pixels, and Kronecker delta function, respectively. However, its descriptive power is limited for cases with a significant imbalance between foreground and background pixels. 
Therefore, mIoU is also utilized in this work as the evaluation metric due to its effectiveness to deal with imbalanced binary and multi-class segmentation. Mean IoU (mIoU) is calculated across classes as Equation \eqref{eqn: miou}: 
\begin{equation}
    mIoU(p,p\prime) = \frac{1}{C}\sum_i^C\frac{\sum_i^N \delta(p_{i,c},1)\delta(p_{i,c},p\prime_{i,c})}{max(1, \delta(p_{i,c},1) + \delta(p\prime_{i,c},1))}
    \label{eqn: miou}
\end{equation}
\noindent where $C$ denotes the number of classes. 
If a pixel $i$ of prediction or ground truth belongs to a certain class c, $p_{i,c}$ and $p\prime_{i,c}$ are 1; otherwise, $p_{i,c}$ and $p\prime_{i,c}$ are 0.

\subsection*{Segmentation on RGB images}\label{sec: RGB}
The approaches for RGBD instance segmentation are sophisticated, so we selected several well-known and widely used methods to evaluate our manually labeled RGB frames, such as FCN \cite{long2015fully}, U-NET\cite{ronneberger2015u}, and DeepLab\cite{chen2017deeplab}. 
The testing results of ESD-1 and ESD-2 datasets using mIoU metrics is 59.36\% on FCN, 64.19\% for U-Net, and 68.77\% for DeepLabV3+. Moreover, the segmentation results on other public conventional datasets MSCOCO \cite{LinMicrosoftContext},PascalVoc\cite{EveringhamTheChallenge}, and CityScape\cite{Cordts2016TheUnderstanding} 
are also listed in Table \ref{tab: evaluation_RGB}. By comparing the testing of known objects ESD-1, the segmentation results of both accuracy and mIoU are lower than most of the public datasets. Since RGB frames of sequences are shuffled and utilized as input for object segmentation, images are blurred when the camera moves. But compared to other datasets with a complex background, our dataset ESD is specific for tabletop objects that is relatively more likely to separate the foreground and the background. For this reason, the segmentation results on RGB images from MSCOCO dataset are relatively low. 
On the other hand, it demonstrates that the RGB part of our dataset is challenging; not only because of the occlusion among objects, but also the impact of the motion blur. In addition, the performance of all testing results on unknown objects from ESD-2 sub-dataset reduces by 30\% approximately.




\begin{table}[ht!]
    \centering
    \begin{tabular}{|c|cc|cc|cc|}
\hline
\multicolumn{1}{|c|}{Datasets} & \multicolumn{2}{c|}{FCN\cite{long2015fully}}                            & \multicolumn{2}{l|}{U-Net\cite{ronneberger2015u}} & \multicolumn{2}{l|}{DeepLab\cite{chen2017deeplab}} \\ \cline{2-7} 
\cline{2-7} 
& \multicolumn{1}{c}{Acc} & \multicolumn{1}{c|}{mIoU} & Acc           & mIoU           & Acc       & mIoU  \\
\hline
ESD Known obj (ours) & 81.37	& 59.36	& 86.27	& 64.19	& 90.59 &	68.77\\
\hline
ESD Unknown obj (ours) & 64.21	&  32.79	&  69.05 	&   40.70	&  72.16  & 43.04	\\
\hline
MSCOCO \cite{LinMicrosoftContext} & 71.6 &	31.43 &	77.2 & 47.21 &	79.13 &	58.01
 \\
 \hline
PascalVoc \cite{EveringhamTheChallenge} & 87.09 & 62.20 &	92.05 &	72.70 &	96.52 &	87.30 
 \\
\hline
Cityscape \cite{Cordts2016TheUnderstanding} & 84.3 & 65.30 & 89.07 &	73.50 & 93.17 & 82.10 
\\
\hline
\end{tabular}
    \caption{Evaluation results of the state-of-the-art segmentation networks FCN, U-Net and DeepLab on RGB frames from ESD. Furthermore, benchmarks of the same networks on other public datasets MSCOCO, PascalVoc and CityScape are also provided. }
    \label{tab: evaluation_RGB}
\end{table}

\subsection*{Segmentation on events data}
As mentioned in \textit{Background} section, there are several approaches for semantic segmentation of autonomous driving, such as EV-SegNet (2019) \cite{alonso2019ev}, VID2E (2019)\cite{gehrig2020video}, EVDistill (2021)\cite{wang2021evdistill}, EV transfer (2022)\cite{messikommer2022bridging}, and ESS (2022)\cite{sun2022ess}. Since there are few deep learning-based approaches for instance segmentation using neuromorphic vision, transfer learning of semantic segmentation can be a possible way to achieve instance segmentation tasks. But some of them are not fully open-sourced nor is the pre-trained model provided. 
As such, it is hard to implement and test these approaches on our datasets. 
Therefore, we employed transfer learning on EV-SegNet and ESS by unfreezing the last 4 convolution layers of the encoder 
the whole decoder module, and the classifier. The testing accuracy is 76.98\% and 81.59\% on EV-SegNet and ESS, respectively. However, mIoU of EV-SegNet and ESS are 7.73\% and 8.92\%, respectively as shown in Table \ref{tab: evaluation_events}. 

\begin{table}[ht!]
\centering
\begin{tabular}{|c|ll|ll|}
\hline
\multicolumn{1}{|c|}{Terms} & \multicolumn{2}{c|}{EV-SegNet \cite{alonso2019ev}}                       & \multicolumn{2}{c|}{ESS\cite{sun2022ess}} \\ \cline{2-5} 

\multicolumn{1}{|c|}{}                        & \multicolumn{1}{c}{Acc} & \multicolumn{1}{c|}{mIoU} & Acc           & mIoU           \\
\hline
ESD  &  76.98 &  7.73& 81.59 & 8.92\\
\hline
\end{tabular}

\caption{Quantitative evaluation results of transfer learning of EV-SegNet \cite{alonso2019ev} and ESS \cite{sun2022ess} on our proposed dataset ESD using accuracy and mIoU. }
\label{tab: evaluation_events}
\end{table}
Compared to the mIoU results on the autonomous driving dataet DDD17 \cite{binas2017ddd17} which is 51.76\% and 51.57\%, the above results demonstrate unsatisfactory segmentation performance on our dataset. This may be due to the significant difference in features between our tabletop objects dataset and the autonomous driving dataset. Our ESD captures events of static objects and backgrounds using moving cameras, which provides homogeneous features on events. However, the DDD17 dataset records dynamic moving objects such as on-road vehicles and pedestrians, providing additional features including various moving velocities, directions, and postures. On the other hand, the comparison indicates that our data is quite challenging due to the similar and homogeneous features of each tabletop object and background. 


\subsection*{Segmentation on integrated RGB and events data}
Since the transfer learning of the event-based semantic segmentation approach fails to provide satisfactory results, we tested our dataset using vision-transformer-based cross-modal fusion networks SA-GATE \cite{chen2020bi} and CMX \cite{liu2022cmx} to extract features from RGB frames and events stream. The quantitative testing results of both known and unknown objects are listed in Table \ref{tab: evaluation_RGB_events}.
For known object segmentation, both models can reach a high accuracy and mIoU of prediction. Compared to the testing results of pure RGB frames, segmentation on integrated RGB and events data can achieve a more accurate segmentation due to the complementary features extracted and fused from both RGB frames and events stream. However, the performance of segmenting unseen objects drops dramatically by 80.66\% and 80.00\% using SA-GATE and CMX, respectively, indicating the challenge of unknown object segmentation. 



\begin{table}[ht!]
    \centering
    \begin{tabular}{|c|ll|ll|}
\hline
\multicolumn{1}{|c|}{Datasets} & \multicolumn{2}{c|}{SA-GATE\cite{chen2020bi}}                            & \multicolumn{2}{l|}{CMX\cite{liu2022cmx}} \\ \cline{2-5} 
\cline{2-5} 

& \multicolumn{1}{c}{Acc} & \multicolumn{1}{c|}{mIoU} & Acc           & mIoU \\
\hline
ESD-1 (Known objects)  & 91.53 	& 84.08	&94.58	& 85.81\\
\hline
ESD-2 (Unknown objects) & 73.04	& 16.26	&  76.78	& 18.90 \\
\hline
\end{tabular}
    \caption{Quantitative evaluation results of SA-GATE \cite{alonso2019ev} and CMX \cite{liu2022cmx} on RGB frames and events stream from our proposed dataset ESD using accuracy and mIoU. }
    \label{tab: evaluation_RGB_events}
\end{table}





\begin{figure*}[!hb]
    \centering
    \subfloat[The case of varying moving trajectories]{\includegraphics[width = 3.3in]{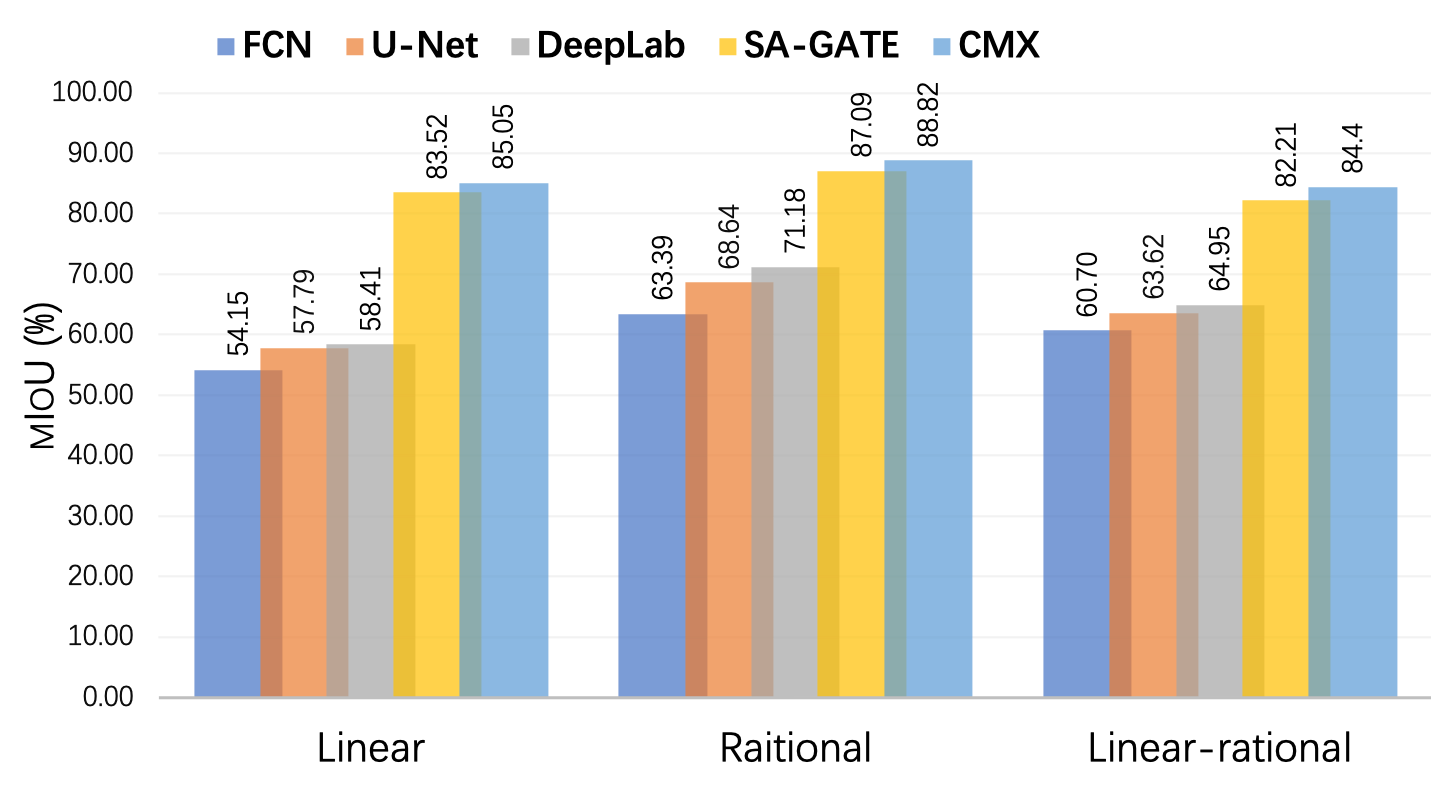}} 
     \subfloat[The case of varying  moving speeds]{\includegraphics[width = 3.3in]{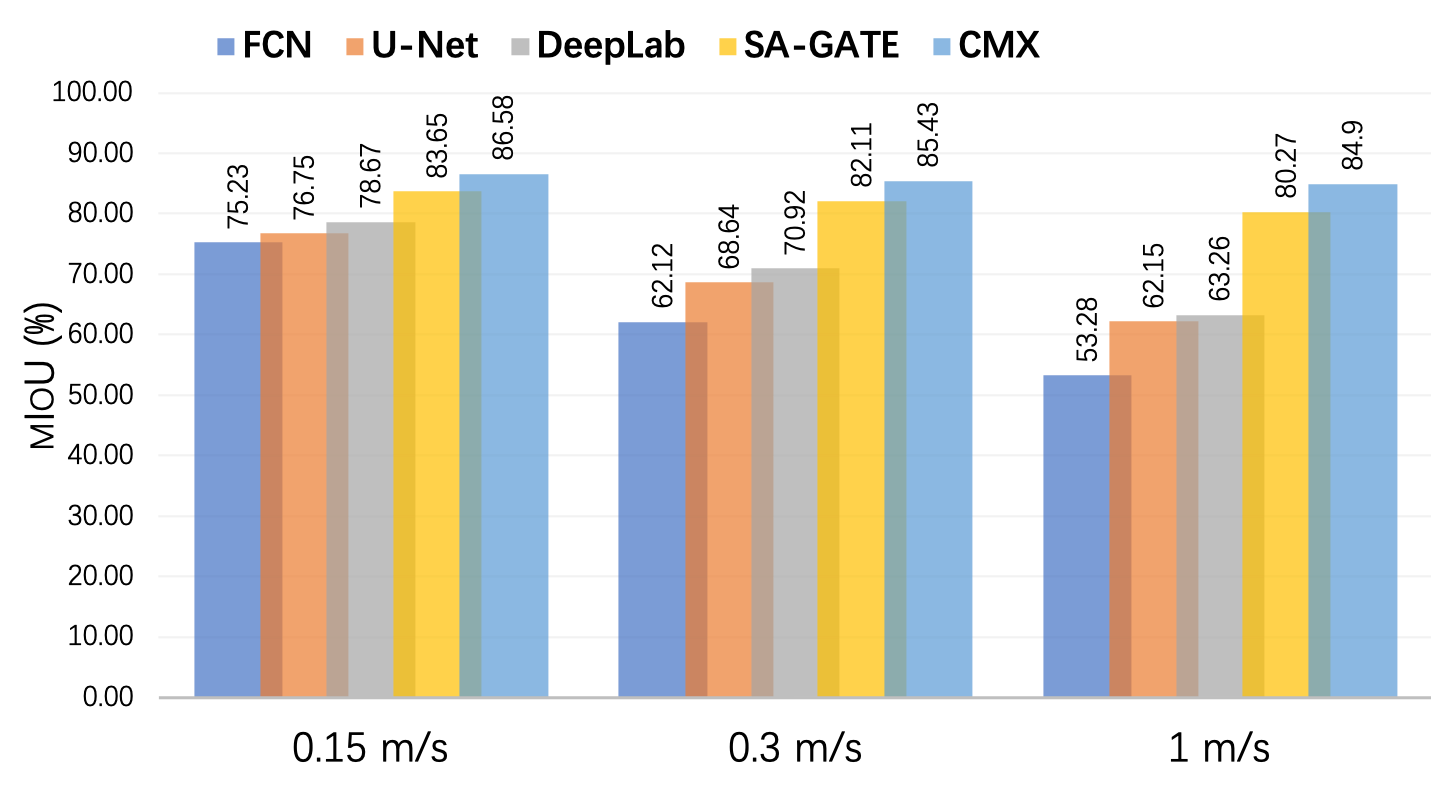}} \\
     \subfloat[The case of varying lighting conditions]{\includegraphics[width = 3.3in]{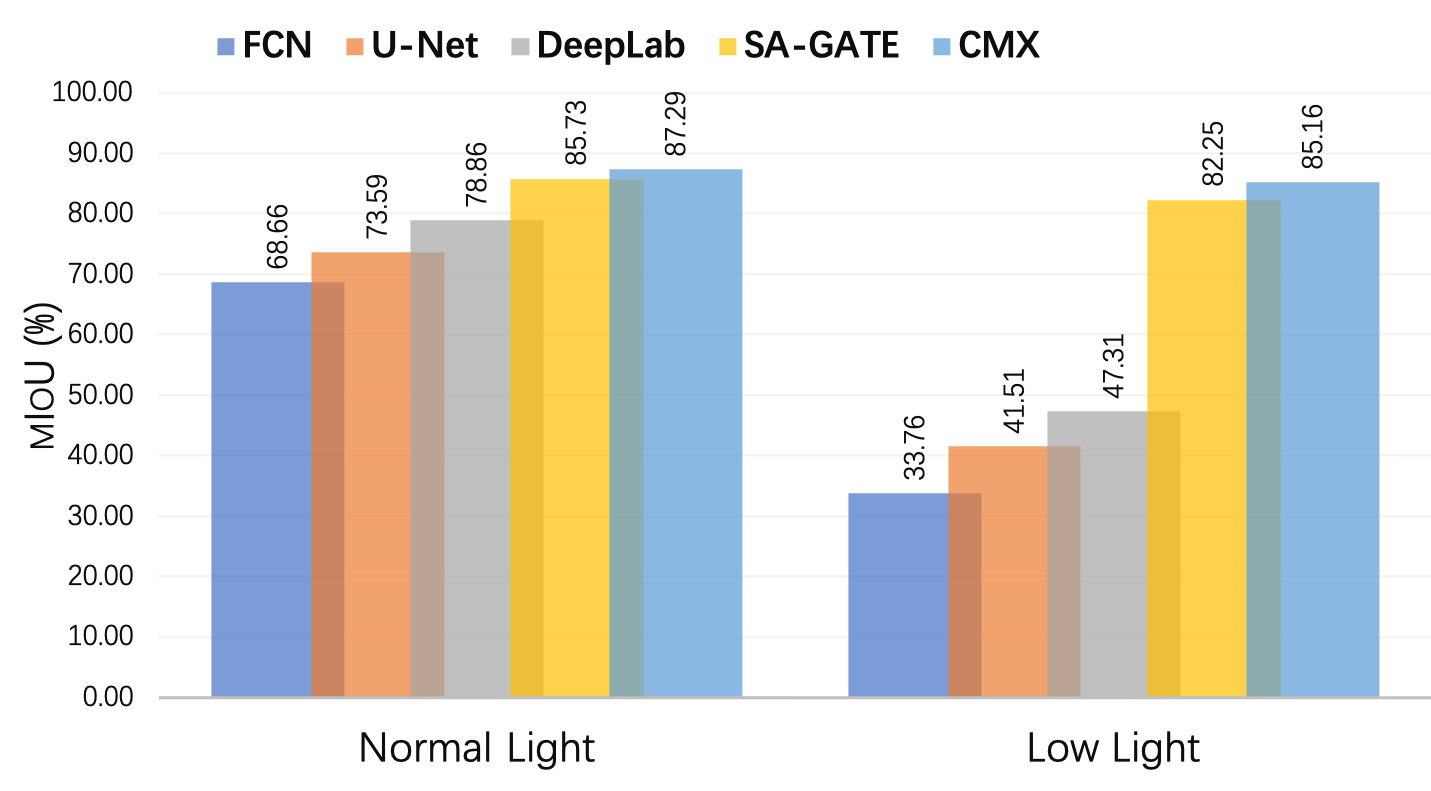}}
    \subfloat[The case of varying heights]{\includegraphics[width = 3.3in]{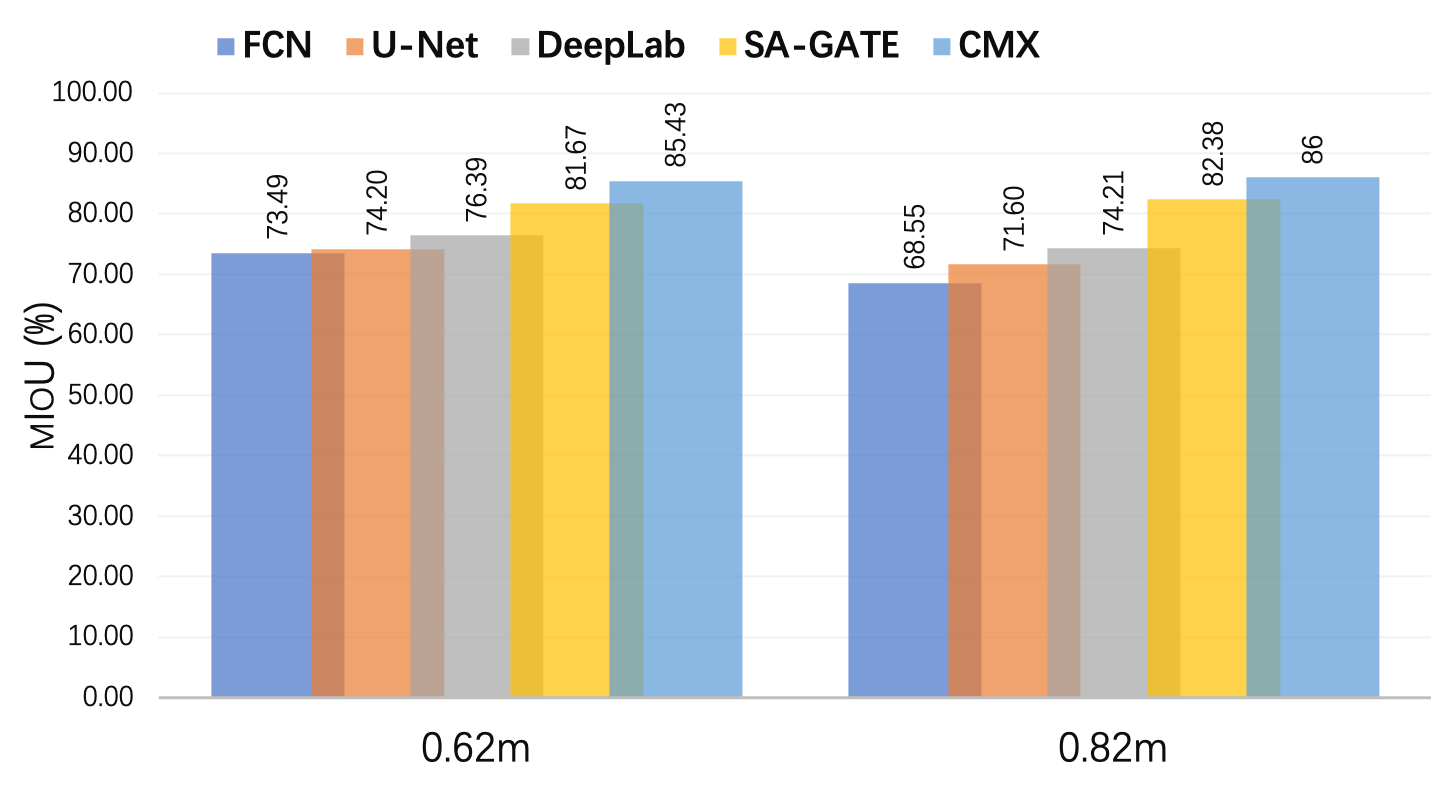}} \\
    \subfloat[The case of varying objects \& w/o occlusion]{\includegraphics[width = 4.5in]{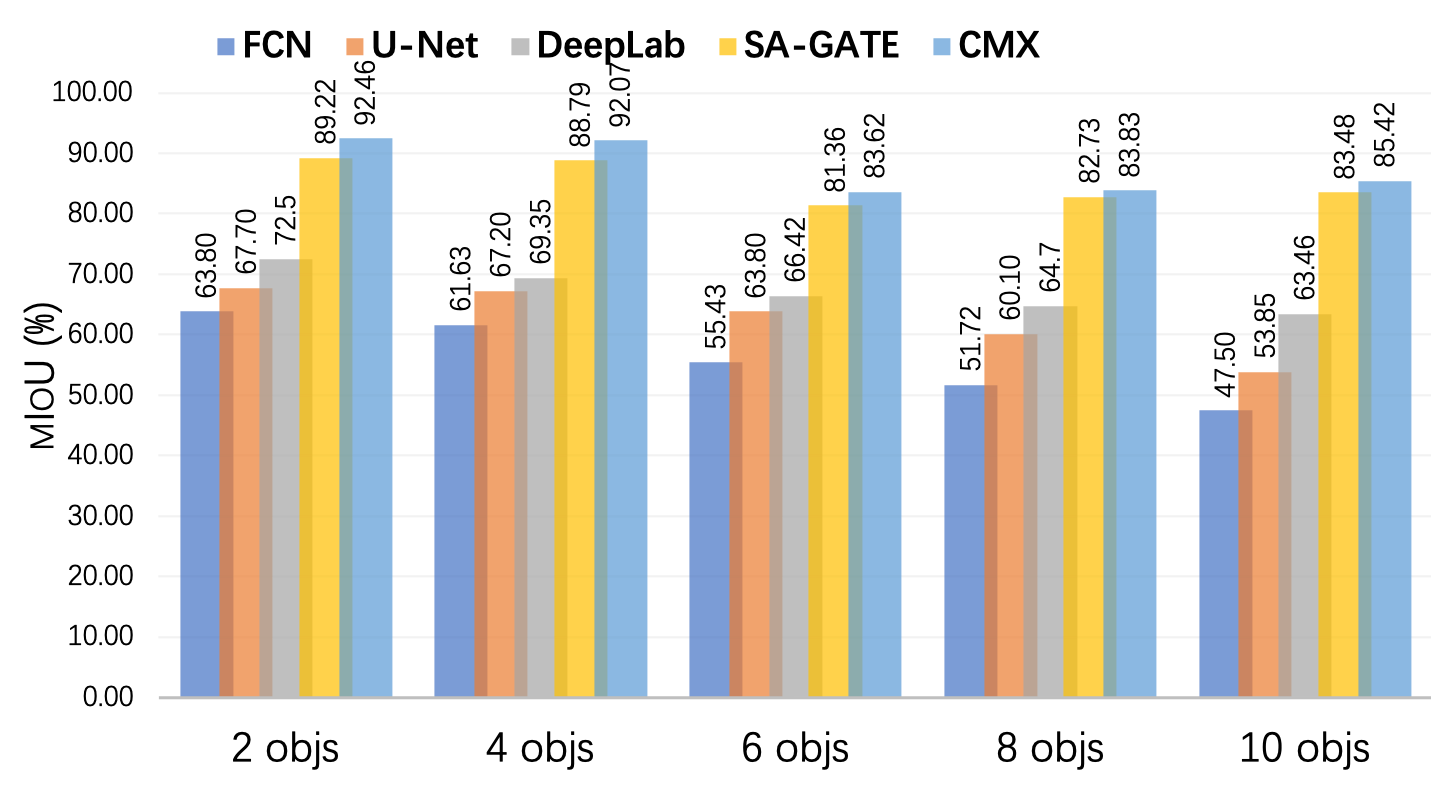}} 
    \caption{The testing results in terms of different attributes: (a) moving trajectories of cameras, (b) moving speed of cameras, (c) lighting condition, (d) the distance between table and cameras, and (e) clutter objects w/o occlusion. Better view in color.}
    \label{fig: results}
\end{figure*}

Besides, the testing results of models utilized in Section \ref{sec: RGB} on RGB frames are also shown in Figure \ref{fig: results} for comparison. 

\paragraph{Varying moving trajectories.} 
We also conducted experiments to compare the performance of the methods according to the type of robotic arm movement or direction of the camera motion. There are three types of robotic arm movement, i.e. linear, rotational, and linear-rotational. In the case of an event-based vision sensor, the direction of motion is an important factor as object edges perpendicular to the motion direction are relatively more exploited than the parallel edges. The impact of the phenomenon can be clearly seen in Figure \ref{fig: results} (a) in terms of the accuracy of segmentation. In general, rotational motion provides rich information as compared to linear motion. Thus, merging the event frames with RGB (CMX model) provides the highest accuracy 88.82\% which is 3.77\%  and 4.42 \% higher than the one of linear and partial linear motion, respectively. 

\paragraph{Moving speed of cameras.}
The testing results under conditions of different moving speeds of cameras are illustrated in Figure \ref{fig: results} (b). Compared to approaches using only RGB frames in Section \ref{sec: RGB}, CMX on both RGB images and events data has the highest mIoU of 85.58\% for 0.15 m/s and it drops to 84.90\% for 1 m/s. The clear impact of event-based vision high speed helps to recover the information at contours and avoid the impact of motion blur in RGB frames. 

\paragraph{Varying lighting conditions.}
Figure \ref{fig: results} (c) demonstrates the testing results under conditions of varying lighting conditions. With the help of event data, our model outperforms other approaches that only use RGB images. Especially in low light conditions, the MIoUs of traditional RGB testing are mostly below 50\%, due to the low perception quality from the conventional images. However, segmentation MIoU reaches around 85\% with the integration of events data, due to the event camera's high sensitivity to the change of light intensity. 

\paragraph{Varying distance between cameras and table.} 
The distance between the camera and the object is varied between $62cm$ and $82cm$, the results are illustrated in Figure \ref{fig: results} (d). Although, there is a minimal impact of the camera and object distance on the accuracy of all the models, yet the effect in the performance of the CMX is  0.67\% compared to the DeepLabV3  2.18\%.

\paragraph{Varying objects/occlusion.}
The segmentation results for different numbers of objects are shown in Figure \ref{fig: results} (e). The scenario of two objects also indicates the condition without occlusion, and scenarios of more than 2 objects represent the occluded condition as depicted in Figure \ref{fig: data_visualization_ESD1}.
Moreover, scenarios are more complex with the increasing number of objects. Thus, 
it can be seen that the mIoU score of all the models for RGB frames keeps decreasing with the increasing number of objects. But the testing results of both cross-modal networks present a U-shape trend, that the lowest value shows in the 6-object scenario because one object is fully stacked on the other object. 





\section*{Code availability}

ALL the events were automatically labeled by the Matlab programs. All Matlab codes are available on GitHub \cite{Xiaoqian_A_Neuromorphic_Dataset_2023} \url{https://github.com/yellow07200/ESD_labeling_tool}. 


\section*{Acknowledgements}

This work was performed at the Advanced Research and Innovation Center (ARIC), which is funded by STRATA Manufacturing PJSC (a Mubadala company), Sandooq Al Watan under Grant SWARD-S22-015, and Khalifa University of Science and Technology.




\bibliography{sample}
\end{document}